\documentclass{article}
\usepackage{amsmath}
\usepackage{algorithm}
\usepackage{algpseudocode}
\usepackage{arxiv}
\usepackage[utf8]{inputenc}
\usepackage[T1]{fontenc}
\usepackage{hyperref}
\usepackage{url}
\usepackage{booktabs}
\usepackage{amsfonts}
\usepackage{nicefrac}
\usepackage{microtype}
\usepackage{lipsum}
\usepackage{graphicx}
\usepackage{tikz}
\usetikzlibrary{positioning, arrows.meta}
\usepackage{tikz}
\usetikzlibrary{positioning, fit, shapes.misc}
\usepackage{subfigure}

\graphicspath{ {./images/} }

\title{Exploring the Potential of Polynomial Basis Functions in Kolmogorov-Arnold Networks: A Comparative Study of Different Groups of Polynomials}

\author{
 Seyd Teymoor Seydi\\
  University of Tehran\\
  Tehran, Iran \\
  \texttt{seydi.teymoor@ut.ac.ir} \\
}

\begin{document}
\maketitle

\begin{abstract}
This paper presents a comprehensive survey of 18 distinct polynomials and their potential applications in Kolmogorov-Arnold Network (KAN) models as an alternative to traditional spline-based methods. The polynomials are classified into various groups based on their mathematical properties, such as orthogonal polynomials, hypergeometric polynomials, q-polynomials, Fibonacci-related polynomials, combinatorial polynomials, and number-theoretic polynomials. The study aims to investigate the suitability of these polynomials as basis functions in KAN models for complex tasks like handwritten digit classification on the MNIST dataset. The performance metrics of the KAN models, including overall accuracy, Kappa, and F1 score, are evaluated and compared. The Gottlieb-KAN model achieves the highest performance across all metrics, suggesting its potential as a suitable choice for the given task. However, further analysis and tuning of these polynomials on more complex datasets are necessary to fully understand their capabilities in KAN models. The source code for the implementation of these KAN models is available at \url{https://github.com/seydi1370/Basis_Functions}.
\end{abstract}

\section{Introduction}
Neural networks have revolutionized various fields, including computer vision, natural language processing, and more. Kolmogorov-Arnold Networks (KANs) \cite{kolmogorov1957representation} have shown promise in approximating complex functions using a combination of simple functions. KANs are inspired by the Kolmogorov-Arnold representation theorem \cite{kolmogorov1957representation}, which states that any continuous function can be represented as a composition of simple functions. 

Recent works have explored the connection between KANs and neural networks. Liu et al. \cite{liu2024kan} proposed a generalized version of KANs that can have arbitrary widths and depths, showing their potential as a promising alternative to multi-layer perceptrons (MLPs). Bozorgasl et al. \cite{bozorgasl2024wavelet} introduced Wav-KAN, which incorporates wavelet functions into the KAN structure to enhance interpretability and performance. Samadi et al. \cite{samadi2024smooth} discussed the importance of smoothness in KANs and proposed smooth, structurally informed KANs that can achieve equivalence to MLPs in specific function classes. SS et al. \cite{ss2024chebyshev} introduced the Chebyshev Kolmogorov-Arnold Network (Chebyshev KAN), which combines the theoretical foundations of the Kolmogorov-Arnold Theorem with the powerful approximation capabilities of Chebyshev polynomials to efficiently approximate nonlinear functions.

KANs offer several advantages over traditional neural network architectures. They provide a more interpretable and efficient representation of complex functions by breaking them down into simpler components \cite{liu2024kan}. KANs can also achieve better accuracy and generalization with fewer parameters compared to MLPs \cite{bozorgasl2024wavelet}. Furthermore, by leveraging inherent structural knowledge, KANs may reduce the data required for training and mitigate the risk of generating hallucinated predictions \cite{samadi2024smooth}.

In this study, we aim to explore the potential of using various polynomials as basis functions in KAN models. We conduct a comprehensive survey of 18 distinct polynomials, classifying them into groups based on their mathematical properties, such as orthogonal polynomials, hypergeometric polynomials, q-polynomials, Fibonacci-related polynomials, combinatorial polynomials, and number-theoretic polynomials. By evaluating the performance of KAN models incorporating these polynomials on the MNIST dataset for handwritten digit classification, we investigate their suitability and effectiveness in approximating complex functions.

The main contributions of this study are as follows:

\begin{enumerate}
    \item A structured overview of 18 polynomials and their characteristics, providing insights into their potential applications in KAN models.
    \item An evaluation of the performance of KAN models incorporating these polynomials on the MNIST dataset. 
    \item An analysis of the relationships between model complexity, number of parameters, training time, and overall performance, offering insights into the factors influencing KAN model performance.
    \item Identification of future research directions, emphasizing the need for further investigation and tuning of these polynomials on more complex datasets and the development of advanced analytical techniques for informed model selection and optimization.
\end{enumerate}




\section{Kolmogorov-Arnold Networks (KANs)} 

\subsection{Kolmogorov-Arnold Representation Theorem}

Kolmogorov-Arnold Networks (KANs) are based on the Kolmogorov-Arnold representation theorem \cite{kolmogorov1957representation}, which states that any continuous multivariate function \( g(z_1, z_2, \dots, z_m) \) of \( m \) variables can be decomposed as:

\[
g(z_1, z_2, \dots, z_m) = \sum_{r=1}^{2m+1} \Phi_r \left( \sum_{s=1}^{m} \phi_{r,s}(z_s) \right),
\]

where:

- \( \phi_{r,s}: [0,1] \to \mathbb{R} \) are inner univariate functions,
- \( \Phi_r: \mathbb{R} \to \mathbb{R} \) are outer univariate functions.

This theorem illustrates that any function of \( m \) variables can be expressed as a sum of compositions of univariate functions, thereby reducing a multivariate problem to simpler univariate transformations.

\subsection{Kolmogorov-Arnold Network (KAN) Formulation}

KANs approximate functions by parameterizing the inner and outer functions with trainable coefficients. 

Let \( \mathbf{y}_p \in \mathbb{R}^{k_p} \) denote the input vector at the \( p \)-th layer of a KAN. The pre-activation value of the \( (p+1, j) \)-th neuron is given by:

\[
y_{p+1,j} = \sum_{i=1}^{k_p} \varrho_{p,i,j}(y_{p,i}),
\]

where:

- \( \varrho_{p,i,j} \) is a learned univariate function applied to the input \( y_{p,i} \),
- \( y_{p+1,j} \) is the sum of all post-activations feeding into the \( (p+1, j) \)-th neuron.

This operation is repeated across all neurons in the layer.

The computation for the entire layer can be written in matrix form as:

\[
\mathbf{y}_{p+1} = \Gamma_p \mathbf{y}_p,
\]

where:

- \( \mathbf{y}_{p+1} \in \mathbb{R}^{k_{p+1}} \) is the vector of outputs from the \( (p+1) \)-th layer,
- \( \mathbf{y}_p \in \mathbb{R}^{k_p} \) is the vector of inputs to the \( p \)-th layer,
- \( \Gamma_p \in \mathbb{R}^{k_{p+1} \times k_p} \) is the function matrix whose elements \( \varrho_{p,i,j} \) are learned functions applied to individual components of \( \mathbf{y}_p \).

Explicitly, \( \Gamma_p \) looks like:

\[
\mathbf{y}_{p+1} = 
\begin{pmatrix}
\varrho_{p,1,1}(y_{p,1}) & \varrho_{p,1,2}(y_{p,2}) & \dots & \varrho_{p,1,k_p}(y_{p,k_p}) \\
\varrho_{p,2,1}(y_{p,1}) & \varrho_{p,2,2}(y_{p,2}) & \dots & \varrho_{p,2,k_p}(y_{p,k_p}) \\
\vdots & \vdots & \ddots & \vdots \\
\varrho_{p,k_{p+1},1}(y_{p,1}) & \varrho_{p,k_{p+1},2}(y_{p,2}) & \dots & \varrho_{p,k_{p+1},k_p}(y_{p,k_p})
\end{pmatrix}
\mathbf{y}_p.
\]

The full KAN model is constructed by composing multiple layers. For a network with \( P \) layers, the output of the KAN model is computed as:

\[
\text{KAN}(\mathbf{y}_0) = \left( \Gamma_{P-1} \circ \Gamma_{P-2} \circ \dots \circ \Gamma_0 \right) \mathbf{y}_0,
\]

where:

- \( \mathbf{y}_0 \in \mathbb{R}^{k_0} \) is the input vector, and
- \( \Gamma_p \) represents the learned transformation at layer \( p \).

If the output dimension \( k_P = 1 \), the final output of the KAN model is a scalar function \( h(\mathbf{y}) \) defined as:

\[
h(\mathbf{y}) \equiv \text{KAN}(\mathbf{y}),
\]

which can be expanded as:

\[
h(\mathbf{y}) = \sum_{i_{P-1}=1}^{k_{P-1}} \varrho_{P-1,1,i_{P-1}} \left( \sum_{i_{P-2}=1}^{k_{P-2}} \dots \left( \sum_{i_1=1}^{k_1} \varrho_{1,i_2,i_1} \left( \sum_{i_0=1}^{k_0} \varrho_{0,i_1,i_0}(y_{i_0}) \right) \right) \right).
\]

Table \ref{tab:polynomial_groups} presents a classification of the 18 polynomials surveyed in the study, along with their respective parameter sets. The polynomials are grouped into several categories based on their mathematical properties and areas of application, such as orthogonal polynomials, hypergeometric polynomials, q-polynomials, Fibonacci-related polynomials, combinatorial polynomials, and number-theoretic polynomials.


\subsubsection{Al-Salam-Carlitz Polynomial}
The Al-Salam-Carlitz polynomials are used in many areas of math and physics \cite{doha2010efficient}. These polynomials are used in the study of quantum algebras and special functions on quantum spaces\cite{klimyk2012quantum}. They are orthogonal polynomials that generalize Chebyshev polynomials\cite{al1965set}.. They have many applications in quantum algebra and the theory of basic hypergeometric series\cite{koekoek1996askey}.

The Al-Salam-Carlitz polynomials $U_n^{(a)}(x;q)$ satisfy the following three-term recurrence relation \cite{al1965set}:
\begin{equation}
\label{eq:recurrence}
U_{n+1}^{(a)}(x;q) = (x-aq^n)U_n^{(a)}(x;q) - q^{n-1}(1-q^n)U_{n-1}^{(a)}(x;q)
\end{equation}

These polynomials form an orthogonal basis with respect to the inner product \cite{ismail2005classical}:

\begin{equation}
\label{eq:inner_product}
\langle f, g \rangle = \int_{-1}^1 f(x)g(x)w(x)dx
\end{equation}

where the weight function \( w(x) \) is given by \cite{koekoek1996askey}:

\begin{equation}
\label{eq:weight_function}
w(x) = \frac{(qx/a, qx/b; q)_\infty}{(x, -qx; q)_\infty}
\end{equation}

Here, \( (a;q)_\infty \) denotes the q-Pochhammer symbol, defined as:

\[
(a;q)_\infty = \prod_{k=0}^{\infty} (1-aq^k)
\]

This symbol plays a crucial role in the context of q-series and orthogonal polynomials \cite{andrews1999special}.

\begin{figure}[htbp]
    \centering
    \includegraphics[width=0.8\textwidth]{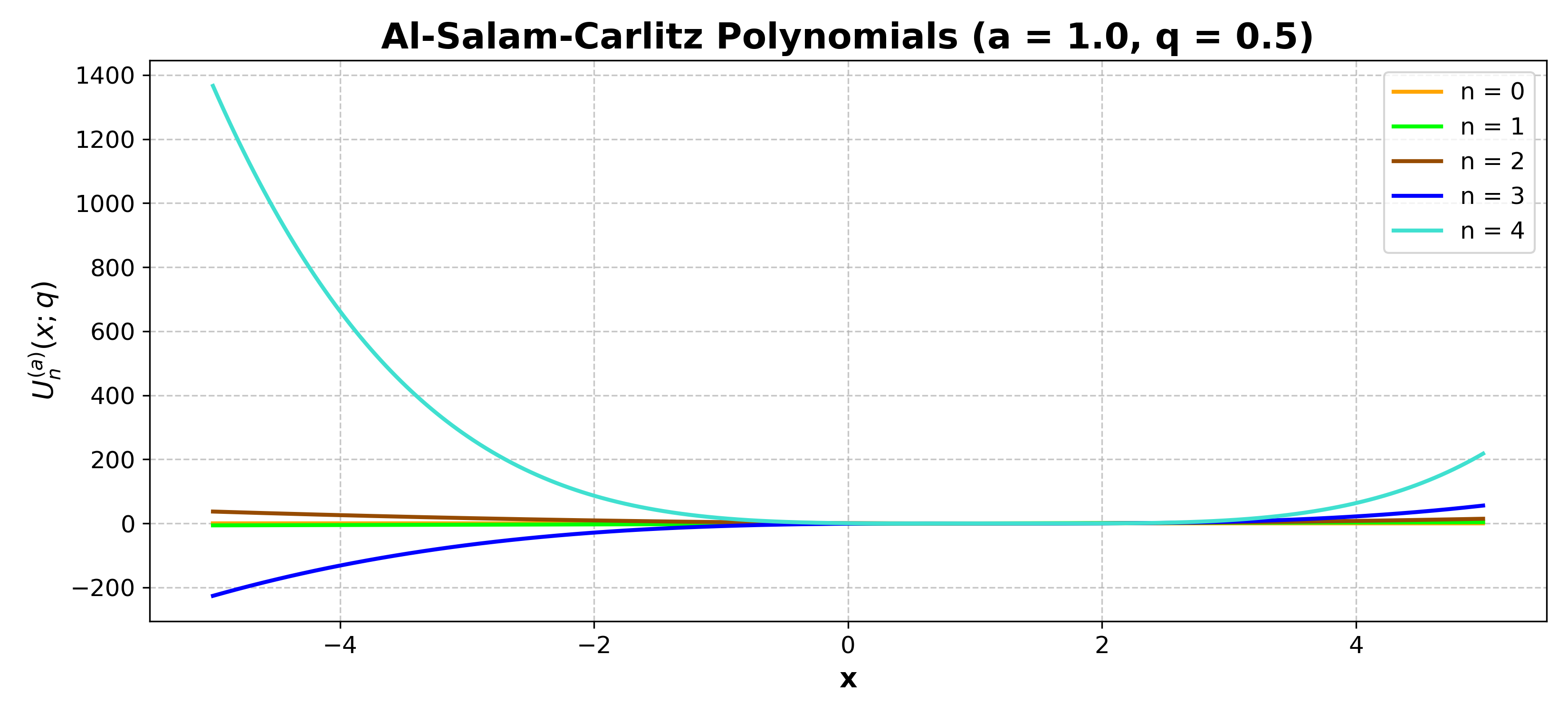}
    \caption{Plot Al-Salam-Carlitz Polynomials with $a = 1.0$ and $q = 0.5$ for various values of $n$.}
    \label{fig:al_salam_carlitz}
\end{figure}

The Figure \ref{fig:al_salam_carlitz} demonstrates the behavior of Al-Salam-Carlitz polynomials for varying degrees (n = 0 to 4) with parameters a = 1.0 and q = 0.5. As the degree n increases, the polynomials exhibit heightened oscillations and accelerated growth, particularly at the edges of the plotted range, thereby illustrating the intricate nature of these special functions.

\subsubsection{Bannai-Ito Polynomial}

Bannai-Ito polynomials constitute a family of orthogonal polynomials that generalize the Racah and Wilson polynomials \cite{bannai1984algebraic}. They play a significant role in the theory of orthogonal polynomials and have deep connections to the Askey scheme of hypergeometric orthogonal polynomials \cite{koekoek1996askey}. These polynomials satisfy the following three-term recurrence relation:

\begin{equation}
\label{eq:bannai_ito_recurrence}
B_{n+1}(x) = (x-\rho_n)B_n(x) - \tau_nB_{n-1}(x)
\end{equation}

where $\rho_n$ and $\tau_n$ are specific coefficient functions that depend on the parameters of the polynomials \cite{tsujimoto2012dunkl}. The Figure \ref{fig:al_salam_carlitz} shows the behavior of the polynomials over the interval \([-5, 5]\). As the degree \(n\) increases, the complexity of the polynomial also increases, leading to more pronounced oscillatory behavior. The colors correspond to different polynomial degrees, ensuring clear visual distinction between the curves. The yellow curve, representing \(n = 4\), exhibits significant growth as \(x\) moves away from the origin, while the lower-degree polynomials remain relatively constrained around zero.

\begin{figure}[htbp]
\centering
\includegraphics[width=0.8\textwidth]{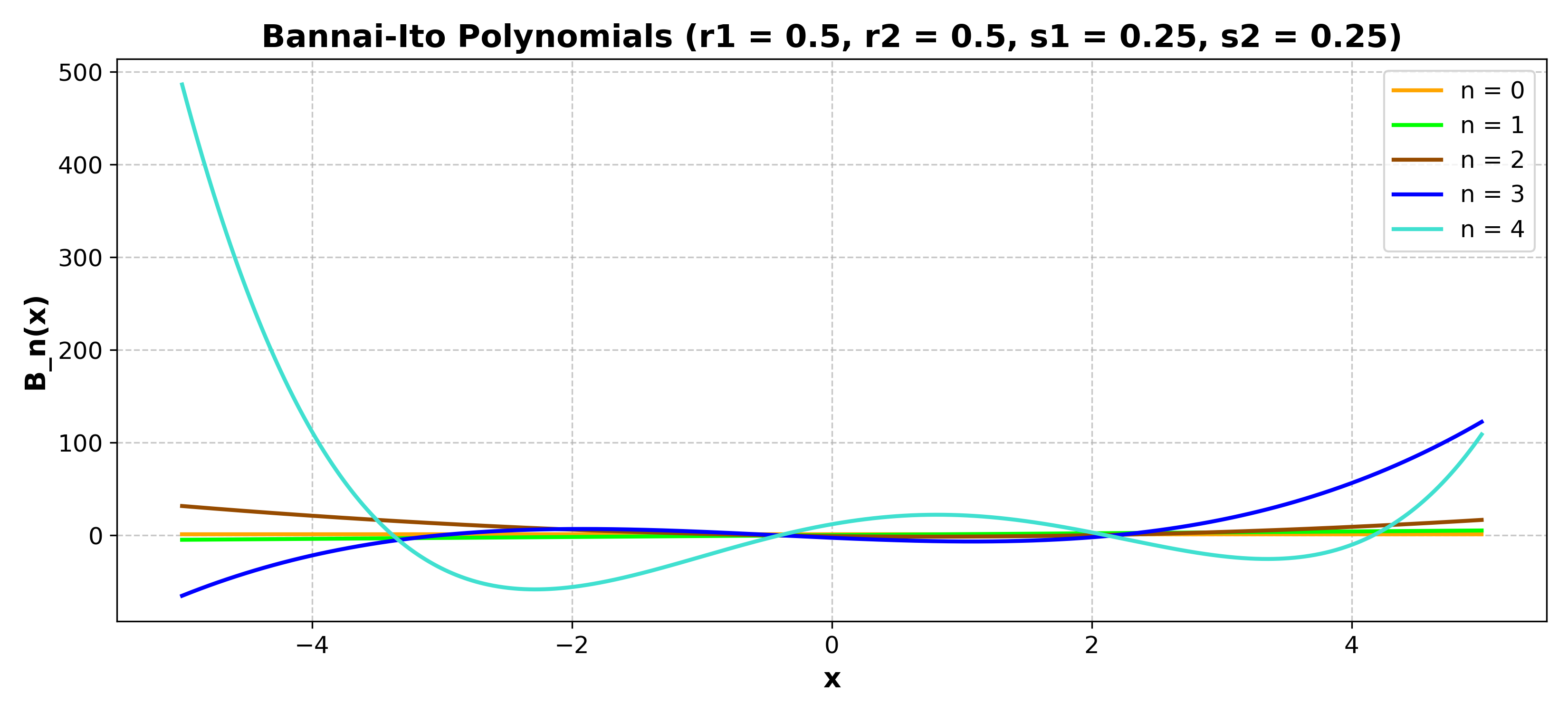}
\caption{Plot of Bannai-Ito Polynomials for various values of $n$ (r1 = 0.5, r2 = 0.5, s1 = 0.25, s2 = 0.25).}
\label{fig:bannai_ito_polynomials}
\end{figure}

The Bannai-Ito polynomials, as illustrated in Figure \ref{fig:bannai_ito_polynomials}, demonstrate increasing oscillation and magnitude with degree $n$. The n=4 polynomial shows significant fluctuations, while lower degrees exhibit more moderate behavior. The polynomials intersect multiple times, with complex behavior especially evident for higher degrees near the interval boundaries.

\subsubsection{Askey-Wilson Polynomial}

Askey-Wilson polynomials constitute a four-parameter family of orthogonal polynomials that generalize numerous classical orthogonal polynomials, including the continuous q-Jacobi polynomials \cite{askey1985some}. ThThey play a pivotal role in the Askey scheme of hypergeometric orthogonal polynomials and have notable applications in quantum groups, special function theory, and mathematical physics \cite{koekoek1996askey}. The Askey-Wilson polynomials can be defined recursively as follows:

\begin{equation}
\label{eq:askey_wilson_recurrence}
p_n(x;a,b,c,d|q) = \frac{(2x - A_n)p_{n-1}(x;a,b,c,d|q) - C_np_{n-2}(x;a,b,c,d|q)}{1-q^n}
\end{equation}

where the initial conditions are:

\begin{equation}
\label{eq:askey_wilson_initial}
\begin{aligned}
p_0(x;a,b,c,d|q) &= 1 \\
p_1(x;a,b,c,d|q) &= \frac{2(1+abq)x - (a+b)(1+cdq)}{1+abcdq^2}
\end{aligned}
\end{equation}

and the coefficients $A_n$ and $C_n$ are given by:

\begin{equation}
\label{eq:askey_wilson_coefficients}
\begin{aligned}
A_n &= \frac{(1-abq^{n-1})(1-cdq^{n-1})(1-abcdq^{2n-2})}{(1-abcdq^{2n-1})(1-abcdq^{2n})} \\
C_n &= \frac{(1-q^n)(1-abq^{n-1})(1-cdq^{n-1})(1-abcdq^{2n-2})}{(1-abcdq^{2n-2})(1-abcdq^{2n-1})}
\end{aligned}
\end{equation}

where, $a$, $b$, $c$, and $d$ are parameters, $q$ is the base of the q-Pochhammer symbol (with $|q| < 1$), and $x = \cos\theta$ \cite{gasper2011basic}.

\begin{figure}[htbp]
\centering
\includegraphics[width=0.8\textwidth]{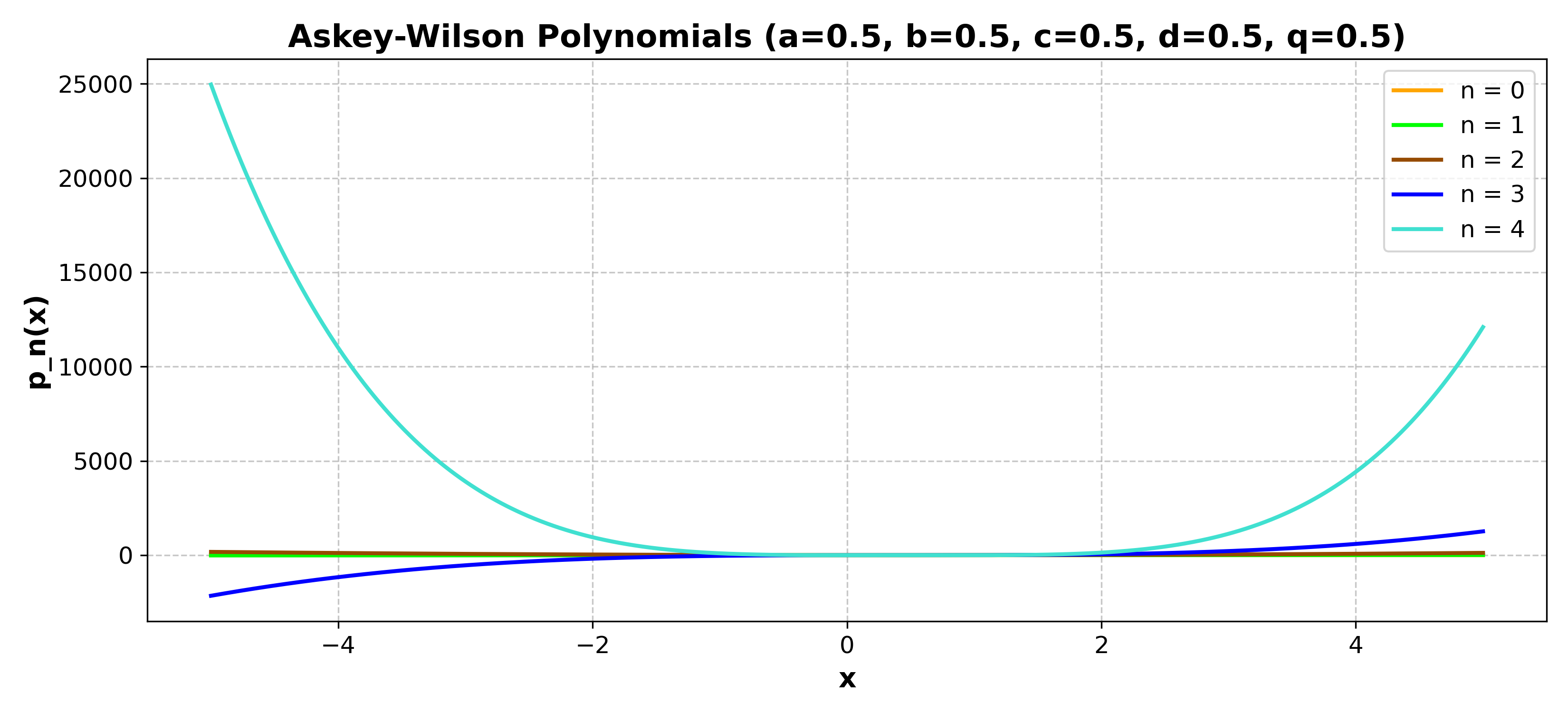}
\caption{Plot of Askey-Wilson Polynomials for various values of $n$ (a=0.5, b=0.5, c=0.5, d=0.5, q=0.5).}
\label{fig:askey_wilson_polynomials}
\end{figure}

The Askey-Wilson polynomials, as shown in Figure \ref{fig:askey_wilson_polynomials}, display increasing complexity with degree $n$. The n=4 polynomial exhibits dramatic growth at interval edges, while lower degrees show more moderate behavior. All polynomials intersect near the origin, with even-degree polynomials appearing symmetric about the y-axis.

\subsubsection{Boas-Buck Polynomial}

Boas-Buck polynomials represent a substantial generalization of several classical orthogonal polynomial families, including Hermite, Laguerre, and Jacobi polynomials \cite{boas1958polynomials}. These polynomials are of great significance in the field of special functions and generating functions, as they offer a unifying framework for the study of diverse polynomial sequences \cite{roman1978umbral}.
The Boas-Buck polynomials $P_n(x)$ satisfy the following three-term recurrence relation:

\begin{equation}
\label{eq:boas_buck_recurrence}
P_n(x) = \frac{(a_{n-1}x - b_{n-1})P_{n-1}(x) - (a_{n-2} - b_{n-2})P_{n-2}(x)}{a_n - b_n}
\end{equation}

with initial conditions:

\begin{equation}
\label{eq:boas_buck_initial}
\begin{aligned}
P_0(x) &= 1 \\
P_1(x) &= \frac{a_0x - b_0}{a_1 - b_1}
\end{aligned}
\end{equation}

where, $\{a_n\}$ and $\{b_n\}$ are sequences of complex numbers that characterize the specific polynomial family within the Boas-Buck class \cite{chihara2011introduction}.
The generating function for Boas-Buck polynomials takes the form:

\begin{equation}
\label{eq:boas_buck_generating}
A(t)e^{xB(t)} = \sum_{n=0}^{\infty} P_n(x)t^n
\end{equation}

where $A(t)$ and $B(t)$ are analytic functions with $A(0) = 1$ and $B'(0) \neq 0$ \cite{andrews1999special}.

\begin{figure}[htbp]
\centering
\includegraphics[width=0.8\textwidth]{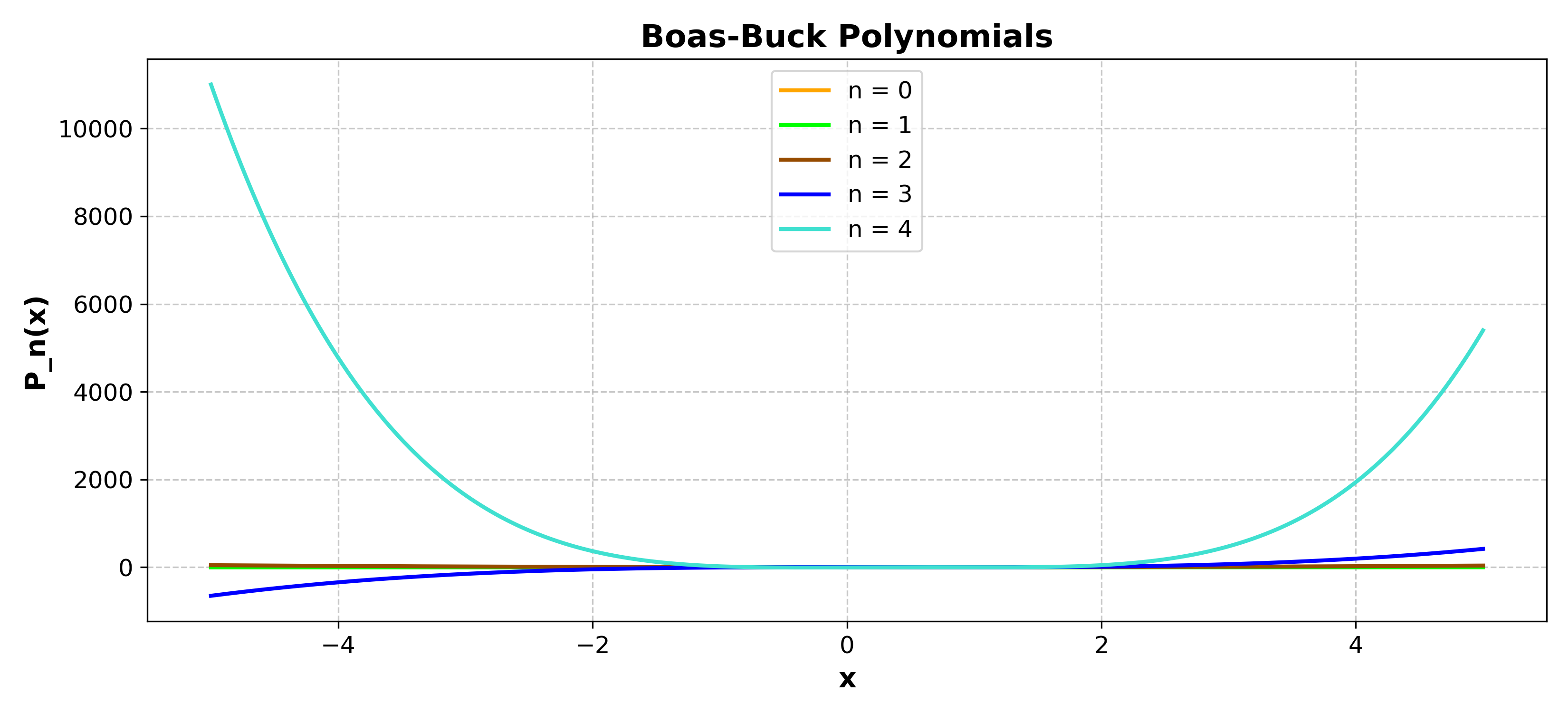}
\caption{Plot of Boas-Buck Polynomials for various values of $n$.}
\label{fig:boas_buck_polynomials}
\end{figure}

The Boas-Buck polynomials, as depicted in Figure \ref{fig:boas_buck_polynomials}, exhibit increasing complexity and magnitude with degree $n$. The n=4 polynomial shows dramatic growth at the interval edges, while lower degrees display more moderate behavior. Even-degree polynomials appear symmetric, and all polynomials intersect near the origin.

\subsubsection{Boubaker Polynomial}

Boubaker polynomials are a family of orthogonal polynomials that have gained attention in mathematical physics, particularly for their applications in heat transfer problems and solutions of certain differential equations \cite{boubaker2007modified}. These polynomials were introduced by Karem Boubaker and have since found applications in various fields of applied mathematics and physics \cite{boubaker2008boubaker}.
The Boubaker polynomials $B_n(x)$ are defined by:

\begin{equation}
\label{eq:boubaker_definition}
B_n(x) = \frac{1}{n!} \frac{d^n}{dx^n} [(1-x^2)^n]
\end{equation}

This definition highlights the polynomial's relationship to the derivatives of $(1-x^2)^n$, which is reminiscent of Rodrigues' formula for classical orthogonal polynomials \cite{chihara2011introduction}.

The Boubaker polynomials satisfy a three-term recurrence relation:

\begin{equation}
\label{eq:boubaker_recurrence}
B_{n+1}(x) = xB_n(x) - n^2B_{n-1}(x)
\end{equation}

with initial conditions $B_0(x) = 1$ and $B_1(x) = x$ \cite{hedi2007sturm}.

\begin{figure}[htbp]
\centering
\includegraphics[width=0.8\textwidth]{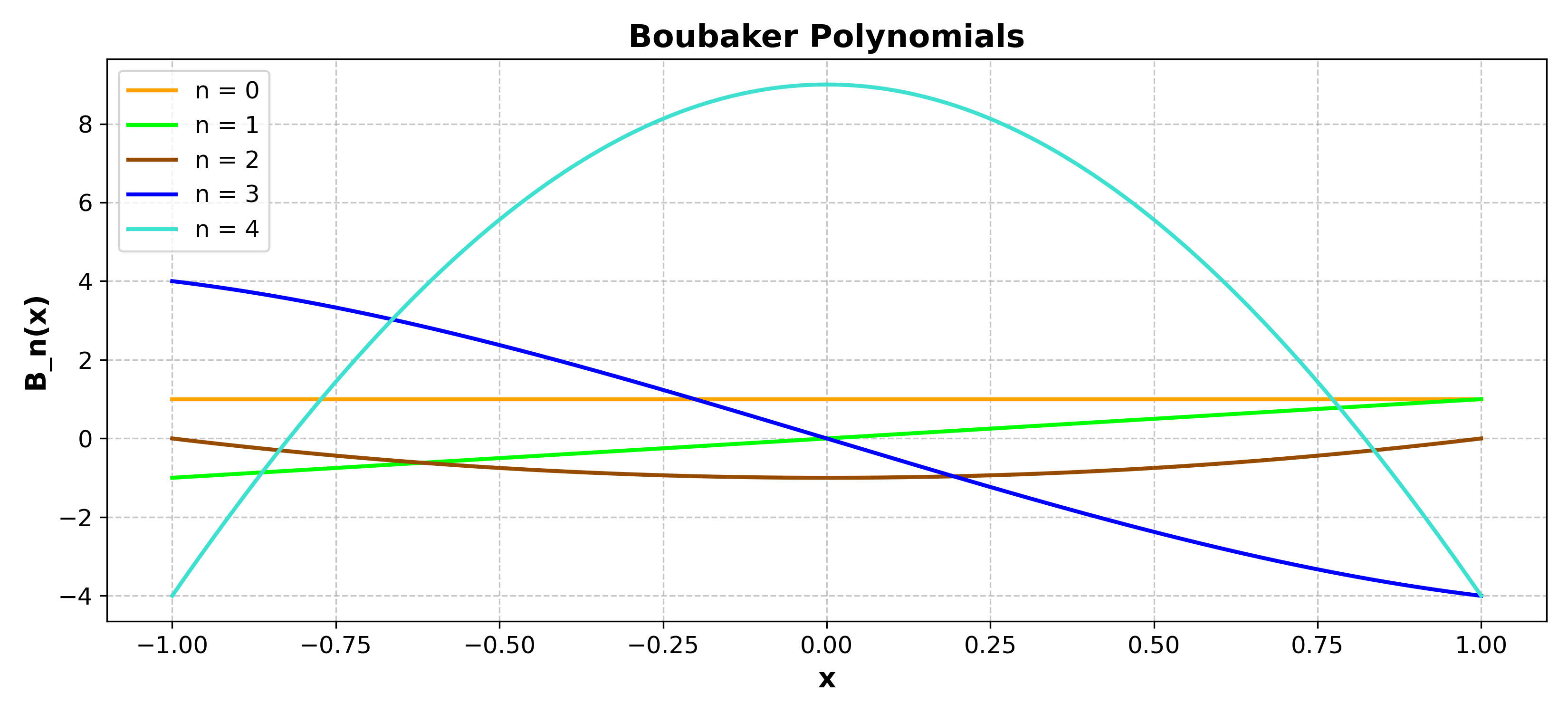}
\caption{Plot of Boubaker Polynomials for various values of $n$.}
\label{fig:boubaker_polynomials}
\end{figure}

The Boubaker polynomials, as shown in Figure \ref{fig:boubaker_polynomials}, display increasing complexity with degree $n$. Even-degree polynomials exhibit symmetry about the y-axis, while odd-degree polynomials show asymmetry. Higher degrees demonstrate more oscillations and larger amplitudes, especially near the interval boundaries.

\subsubsection{Fermat Polynomial}
Fermat polynomials represent a fascinating family of polynomials that play a significant role in number theory and have intriguing connections to Fermat's Last Theorem \cite{ribenboim1995fermat}. These polynomials display a multitude of intricate algebraic characteristics and are intimately associated with cyclotomic polynomials\cite{washington2012introduction}.
The Fermat polynomial $F_n(x)$ of degree $n$ is defined as:

\begin{equation}
\label{eq:fermat_definition}
F_n(x) = \prod_{k=0}^{n-1} (x - e^{2\pi ik/n})
\end{equation}

This definition reveals that the roots of $F_n(x)$ are the $n$-th roots of unity, excluding 1 \cite{lang2012algebra}.

\begin{figure}[htbp]
    \centering
    \includegraphics[width=0.8\textwidth]{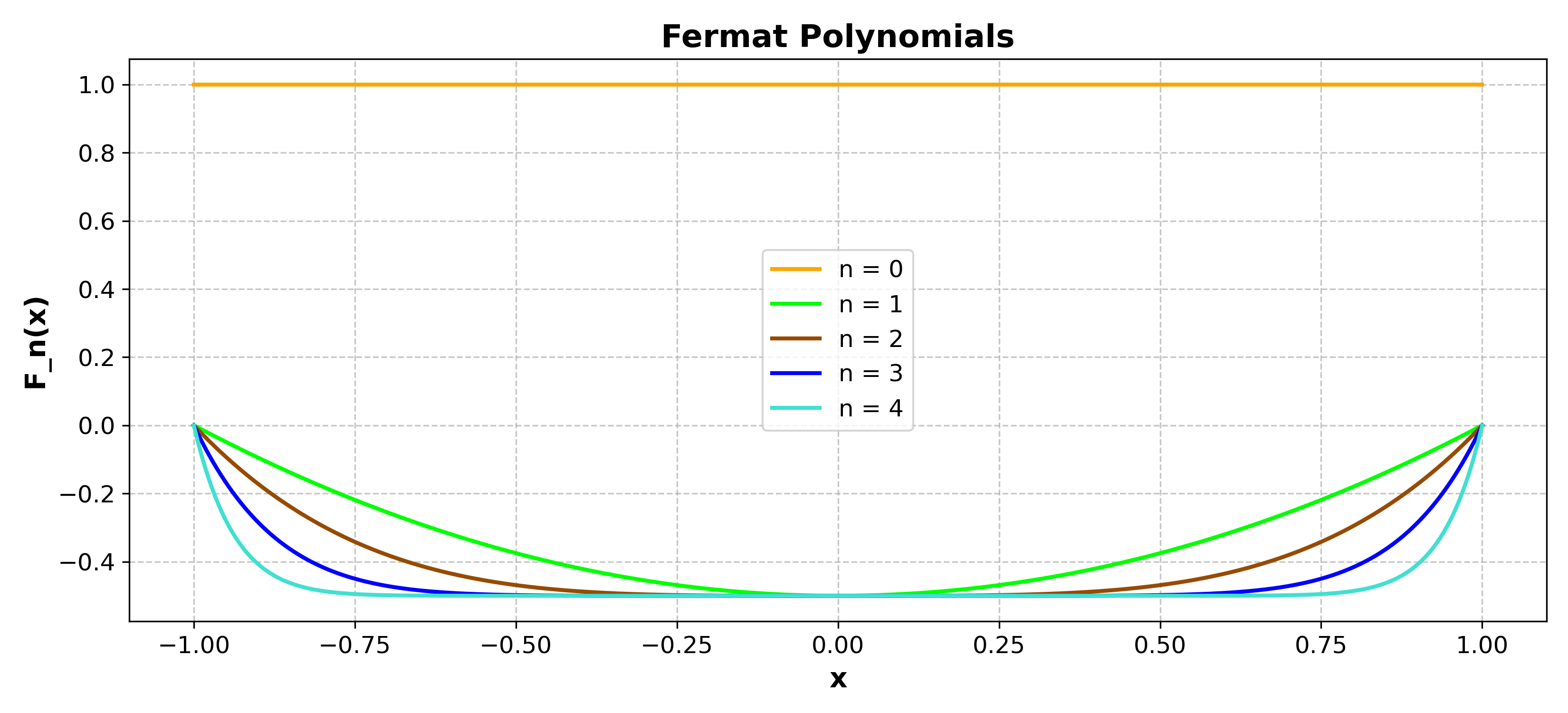}
    \caption{Plot of Fermat Polynomials for various values of $n$.}
    \label{fig:fermat_polynomials}
\end{figure}

The Fermat polynomials, as illustrated in Figure \ref{fig:fermat_polynomials}, exhibit increasing oscillation as the degree $n$ rises from 0 to 4. The polynomials have $n$ distinct real zeros and display symmetric behavior around the y-axis. As $n$ increases, the polynomials demonstrate more significant fluctuations, particularly near the interval's boundaries, underlining the intricate properties of these polynomials for higher degrees.

\subsubsection{Gottlieb Polynomial}

The Gottlieb polynomials, first introduced by D. H. Gottlieb in 1938, constitute a fascinating family of polynomials with notable connections to the Bernoulli numbers. They have found applications in diverse fields, including combinatorics and number theory \cite{gottlieb1938concerning}. These polynomials serve as a conduit between discrete and continuous mathematics, shedding light on the behavior of Bernoulli numbers within a polynomial framework.

The Gottlieb polynomial $G_n(x)$ of degree $n$ is defined as:

\begin{equation}
\label{eq:gottlieb_definition}
G_n(x) = \sum_{k=0}^n \binom{n}{k} B_k x^{n-k}
\end{equation}

where $B_k$ are the Bernoulli numbers \cite{roman1978umbral}.

\begin{figure}[htbp]
    \centering
    \includegraphics[width=0.8\textwidth]{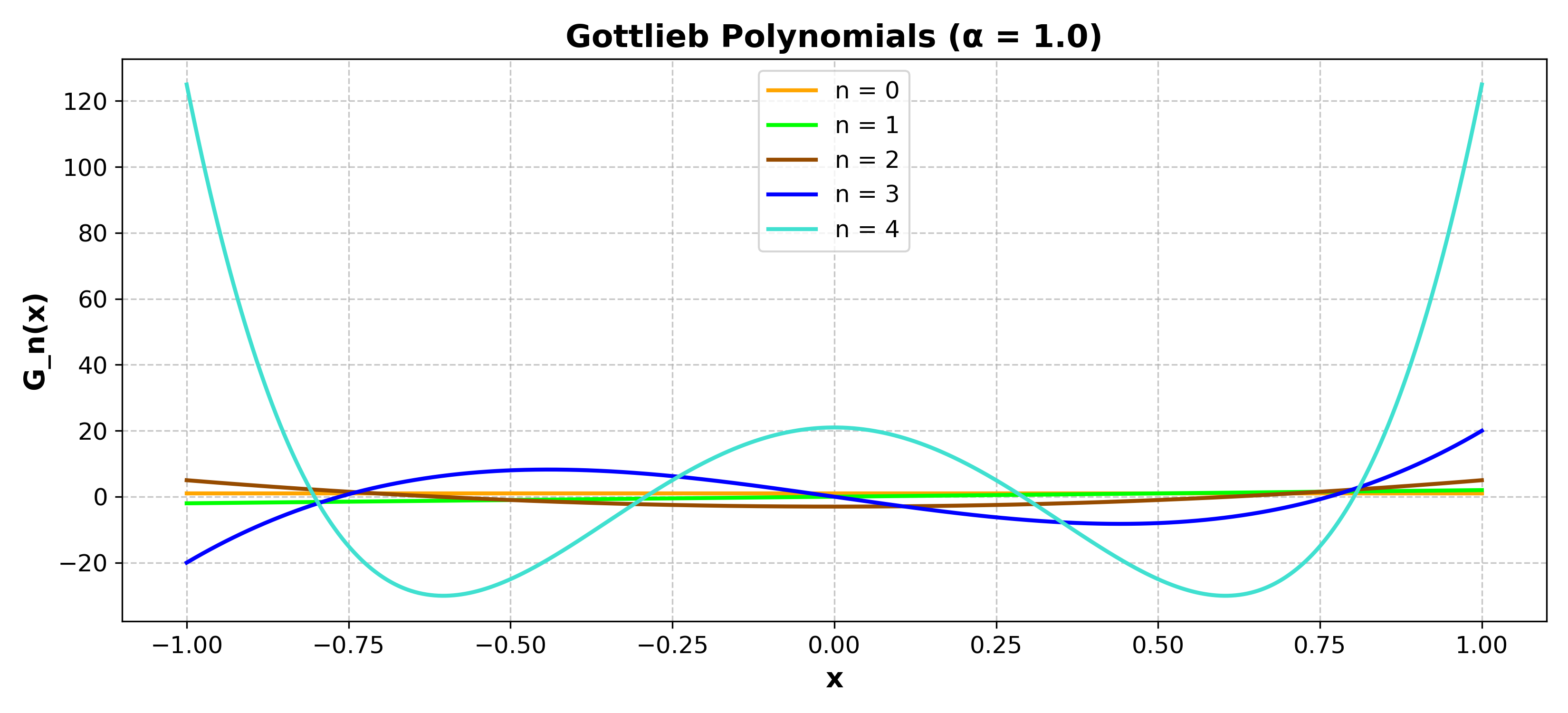}
    \caption{Plot of Gottlieb Polynomials ($\alpha = 1.0$) for various values of $n$.}
    \label{fig:gottlieb_polynomials}
\end{figure}

The Gottlieb polynomials, parameterized by $\alpha = 1.0$, demonstrate increasing oscillation and rapid growth as the degree $n$ rises from 0 to 4, as shown in Figure \ref{fig:gottlieb_polynomials}. The polynomials exhibit $n$ real, distinct zeros and display more significant fluctuations, particularly near the interval's endpoints, emphasizing the complex behavior of these polynomials for higher degrees.

\subsubsection{Heptanacci Polynomial}
Heptanacci polynomials represent a fascinating generalization of both the Fibonacci polynomials and the heptanacci numbers. These polynomials constitute a subclass of generalized Fibonacci polynomials and have notable applications in combinatorics and number theory \cite{kilic2008binet}. The examination of these polynomials offers insights into the domain of recurrence relations and their polynomial analogues.

The Heptanacci polynomials $H_n(x)$ are defined by the following recurrence relation:

\begin{equation}
\label{eq:heptanacci_recurrence}
H_n(x) = xH_{n-1}(x) + H_{n-2}(x) + H_{n-3}(x) + H_{n-4}(x) + H_{n-5}(x) + H_{n-6}(x) + H_{n-7}(x)
\end{equation}

for $n \geq 7$, with initial conditions:
\begin{equation}
\label{eq:heptanacci_initial}
\begin{aligned}
H_0(x) &= 0, \quad H_1(x) = 1, \quad H_2(x) = x, \quad H_3(x) = x^2 + 1, \\
H_4(x) &= x^3 + 2x, \quad H_5(x) = x^4 + 3x^2 + 1, \quad H_6(x) = x^5 + 4x^3 + 3x
\end{aligned}
\end{equation}

\begin{figure}[htbp]
    \centering
    \includegraphics[width=0.8\textwidth]{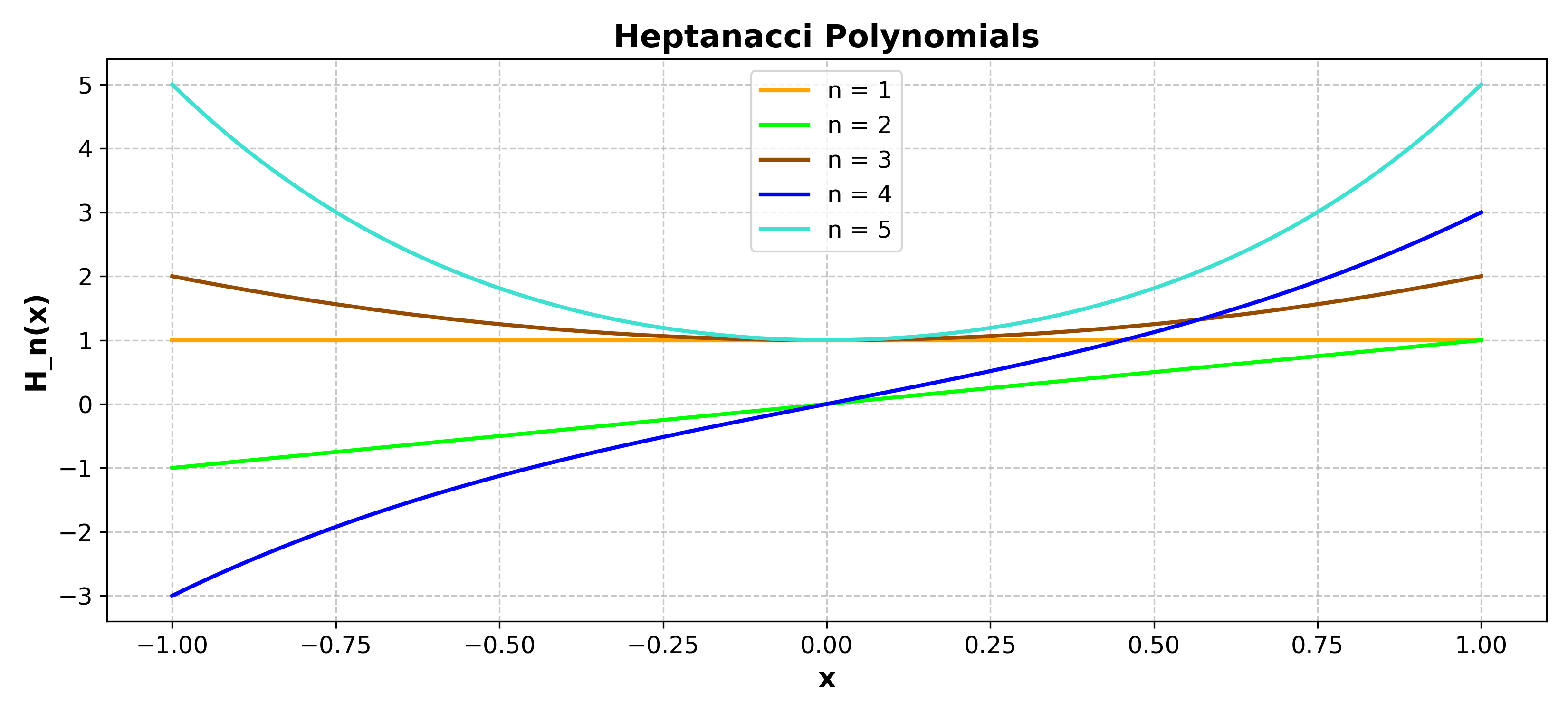}
    \caption{Plot of Heptanacci Polynomials for various values of $n$.}
    \label{fig:heptanacci_polynomials}
\end{figure}

The Heptanacci polynomials, depicted in Figure \ref{fig:heptanacci_polynomials}, exhibit increasing oscillation and more pronounced slopes as the degree $n$ increases from 1 to 5. The polynomials have $n$ real zeros and display greater fluctuations, especially near the edges of the plotted interval, underlining the intricate nature of these polynomials for higher degrees.

\subsubsection{Hexanacci Polynomial}
Hexanacci polynomials represent an intriguing generalization of both the Fibonacci polynomials and the hexanacci numbers. These polynomials belong to a broader class of generalized Fibonacci polynomials and have significant applications in combinatorics and number theory \cite{kilic2007generalized}. The study of hexanacci polynomials offers valuable insights into higher-order recurrence relations and their polynomial analogues.
The Hexanacci polynomials $H_n(x)$ are defined by the following recurrence relation:

\begin{equation}
\label{eq:hexanacci_recurrence}
H_n(x) = xH_{n-1}(x) + H_{n-2}(x) + H_{n-3}(x) + H_{n-4}(x) + H_{n-5}(x) + H_{n-6}(x)
\end{equation}

for $n \geq 6$, with initial conditions:
\begin{equation}
\label{eq:hexanacci_initial}
\begin{aligned}
H_0(x) &= 0, \quad H_1(x) = 1, \quad H_2(x) = x, \quad H_3(x) = x^2 + 1, \\
H_4(x) &= x^3 + 2x, \quad H_5(x) = x^4 + 3x^2 + 1
\end{aligned}
\end{equation}

\begin{figure}[htbp]
    \centering
    \includegraphics[width=0.8\textwidth]{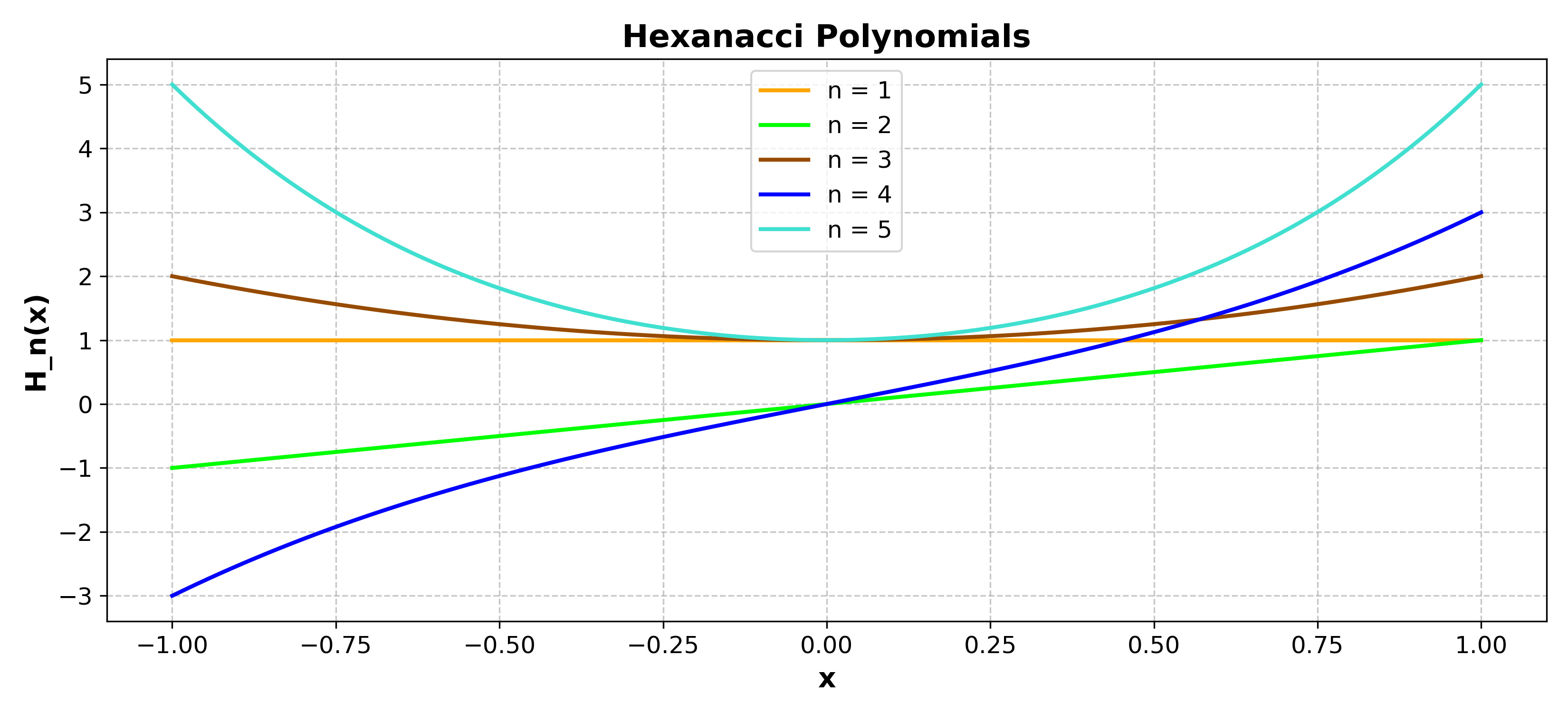}
    \caption{Plot of Hexanacci Polynomials for various values of $n$.}
    \label{fig:hexanacci_polynomials}
\end{figure}

The Hexanacci polynomials, as shown in Figure \ref{fig:hexanacci_polynomials}, display increasing oscillation and steeper slopes as the degree $n$ increases from 1 to 5. The polynomials intersect the x-axis at $n$ distinct points and exhibit more pronounced fluctuations near the interval's boundaries, highlighting the polynomials' complex behavior for higher degrees.

\subsubsection{Meixner-Pollaczek Polynomial}

The Meixner-Pollaczek polynomials constitute a notable family of orthogonal polynomials that generalize the Laguerre polynomials. These polynomials, initially introduced by Meixner and subsequently elaborated upon by Pollaczek, occupy a pivotal position in the realms of quantum mechanics and the theory of special functions \cite{pollaczek1950famille}. They provide a continuous interpolation between Laguerre and Meixner polynomials, offering a versatile tool in mathematical physics and analysis.
The Meixner-Pollaczek polynomials $P_n^{(\lambda)}(x;\phi)$ are defined by:

\begin{equation}
\label{eq:meixner_pollaczek_definition}
P_n^{(\lambda)}(x;\phi) = \frac{(2\lambda)_n}{n!} e^{in\phi} {_2F_1}\left(-n,\lambda+ix;\, 2\lambda;\, 1-e^{-2i\phi}\right)
\end{equation}

where $\lambda > 0$, $0 < \phi < \pi$, $(a)_n$ is the Pochhammer symbol, and $_2F_1$ is the Gaussian hypergeometric function \cite{koekoek1996askey}.

\begin{figure}[htbp]
    \centering
    \includegraphics[width=0.8\textwidth]{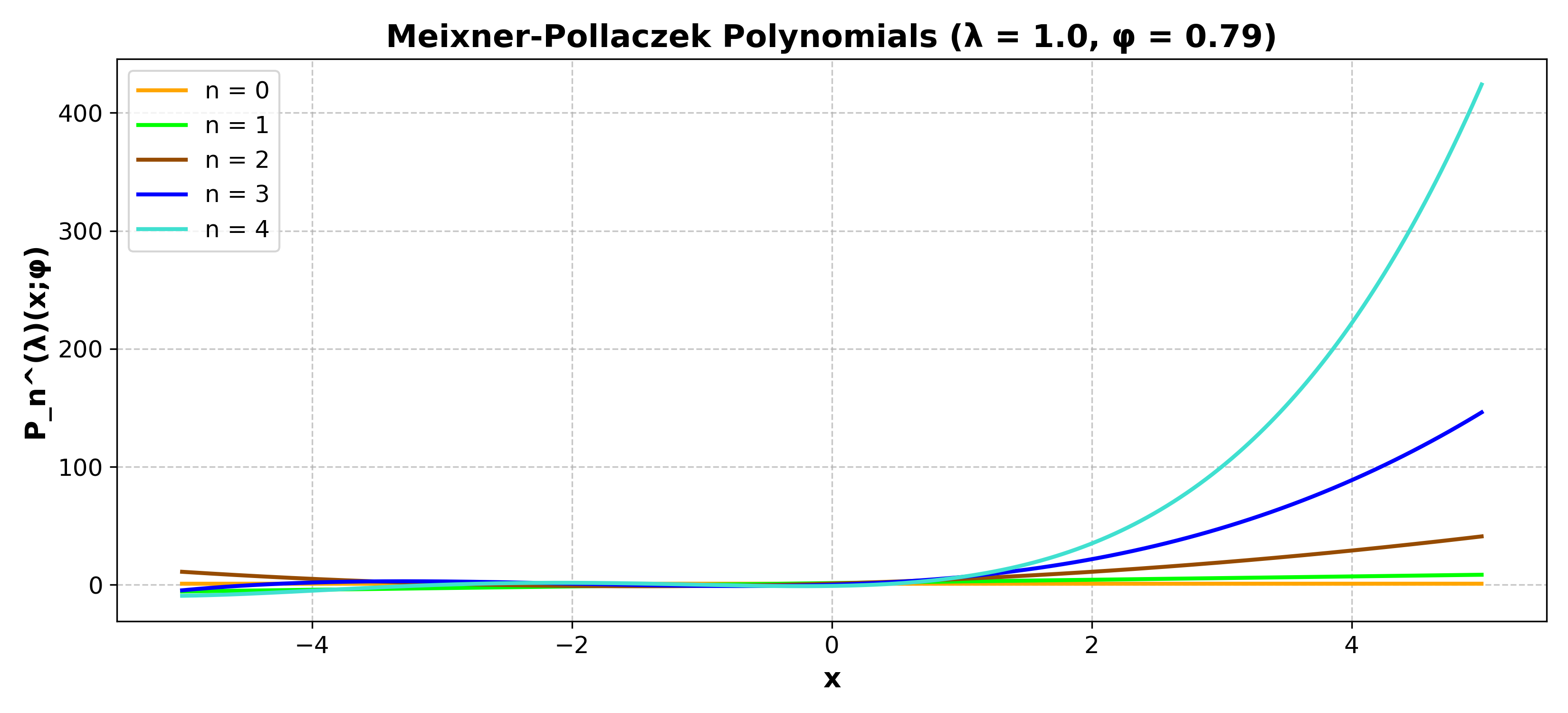}
    \caption{Plot of Meixner-Pollaczek Polynomials ($\lambda = 1.0$, $\phi = 0.79$) for various values of $n$.}
    \label{fig:meixner_pollaczek_polynomials}
\end{figure}

The Meixner-Pollaczek polynomials, parameterized by $\lambda = 1.0$ and $\phi = 0.79$, exhibit increasing oscillation and magnitude as the degree $n$ rises from 0 to 4, as illustrated in Figure \ref{fig:meixner_pollaczek_polynomials}. As $n$ increases, the polynomials have $n$ real, distinct zeros and demonstrate greater fluctuations, particularly near the interval's endpoints, showcasing the polynomials' intricate behavior.

\subsubsection{Narayana Polynomial}
\begin{figure}[htbp]
    \centering
    \includegraphics[width=0.8\textwidth]{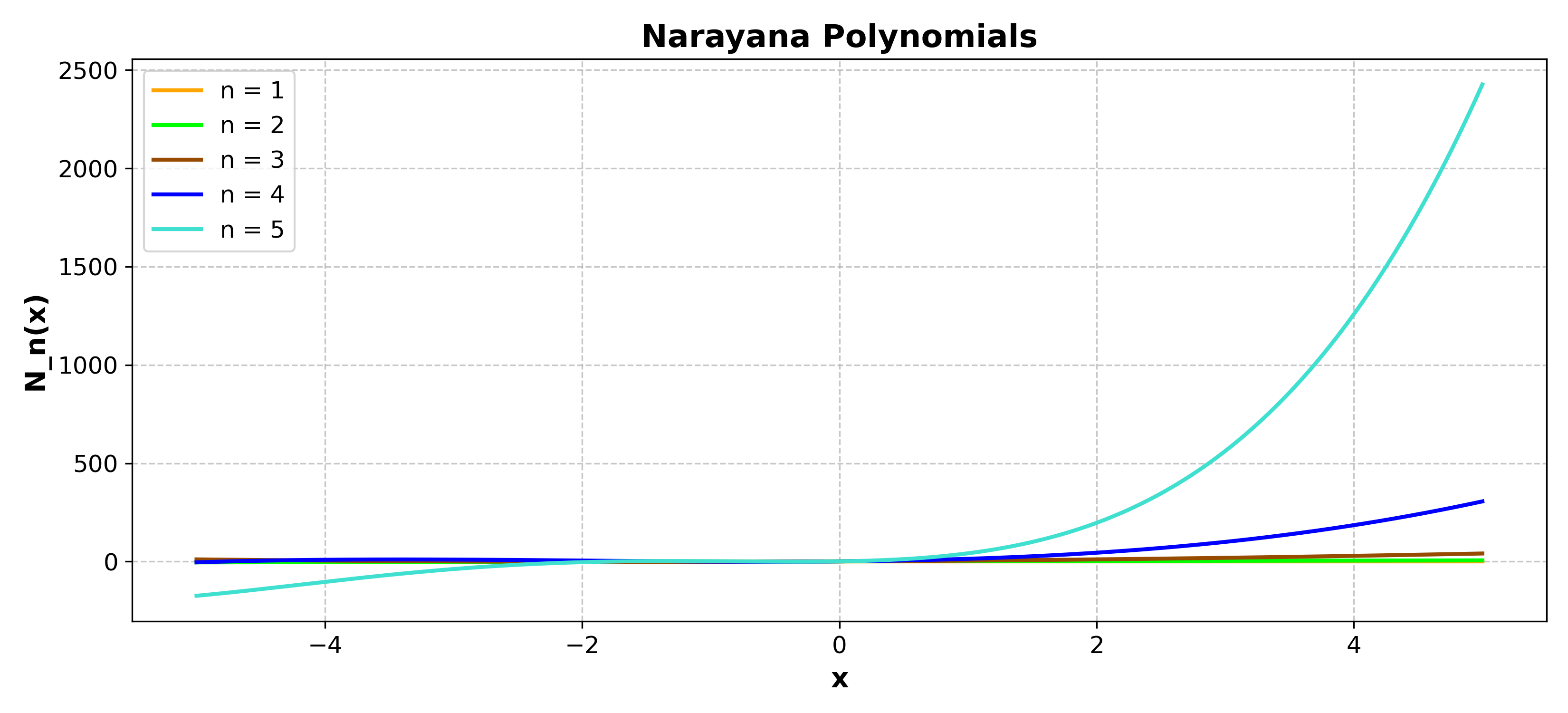}
    \caption{Plot of Narayana Polynomials for various values of $n$.}
    \label{fig:narayana_polynomials}
\end{figure}

Narayana polynomials constitute an intriguing family of polynomials with profound connections to combinatorics, particularly in lattice path enumeration and the theory of Dyck words \cite{narayana1979lattice}. These polynomials offer a refined enumeration of specific combinatorial objects and are intimately connected to the renowned Catalan numbers.
The Narayana polynomial $N_n(x)$ of degree $n-1$ is defined as:

\begin{equation}
\label{eq:narayana_definition}
N_n(x) = \sum_{k=1}^n N(n,k) x^{k-1} = \sum_{k=1}^n \frac{1}{n} \binom{n}{k} \binom{n}{k-1} x^{k-1}
\end{equation}

where $N(n,k)$ is the Narayana number, which counts various combinatorial objects \cite{petersen2015eulerian}.
Figure \ref{fig:narayana_polynomials} demonstrates the behavior of Narayana polynomials for degrees n = 1 to 5. As the degree of the polynomial increases, the value of the polynomial displays a rapid growth, particularly towards the right end of the plotted interval. Additionally, the higher-degree polynomials demonstrate a more pronounced curvature.

\subsubsection{Octanacci Polynomial}

Octanacci polynomials represent an intriguing generalization of both the Fibonacci polynomials and the octanacci numbers. These polynomials are members of a broader class of generalized Fibonacci polynomials. The study of Octanacci polynomials offers significant insights into higher-order recurrence relations and their polynomial analogues.

The Octanacci polynomials $O_n(x)$ are defined by the following recurrence relation:

\begin{equation}
\label{eq:octanacci_recurrence}
O_n(x) = xO_{n-1}(x) + O_{n-2}(x) + O_{n-3}(x) + O_{n-4}(x) + O_{n-5}(x) + O_{n-6}(x) + O_{n-7}(x)
\end{equation}

for $n \geq 7$, with initial conditions:

\begin{equation}
\label{eq:octanacci_initial}
\begin{aligned}
O_0(x) &= 0, \quad O_1(x) = 1, \quad O_2(x) = x, \quad O_3(x) = x^2 + 1, \\
O_4(x) &= x^3 + 2x, \quad O_5(x) = x^4 + 3x^2 + 1, \quad O_6(x) = x^5 + 4x^3 + 3x
\end{aligned}
\end{equation}

The Octanacci numbers, which are obtained when $x = 1$ in the Octanacci polynomials, form the sequence \cite{oeis_octanacci}:

0, 1, 1, 2, 4, 8, 16, 32, 64, 128, 255, 509, 1016, 2028, 4048, 8080, 16128, 32192, 64255, ...

These numbers satisfy the recurrence relation:

\begin{equation}
\label{eq:octanacci_number_recurrence}
O_n = O_{n-1} + O_{n-2} + O_{n-3} + O_{n-4} + O_{n-5} + O_{n-6} + O_{n-7}
\end{equation}

for $n \geq 7$, with initial conditions $O_0 = 0$, $O_1 = O_2 = 1$, and $O_3 = O_4 = O_5 = O_6 = 2^{n-3}$.

\begin{figure}[htbp]
    \centering
    \includegraphics[width=0.8\textwidth]{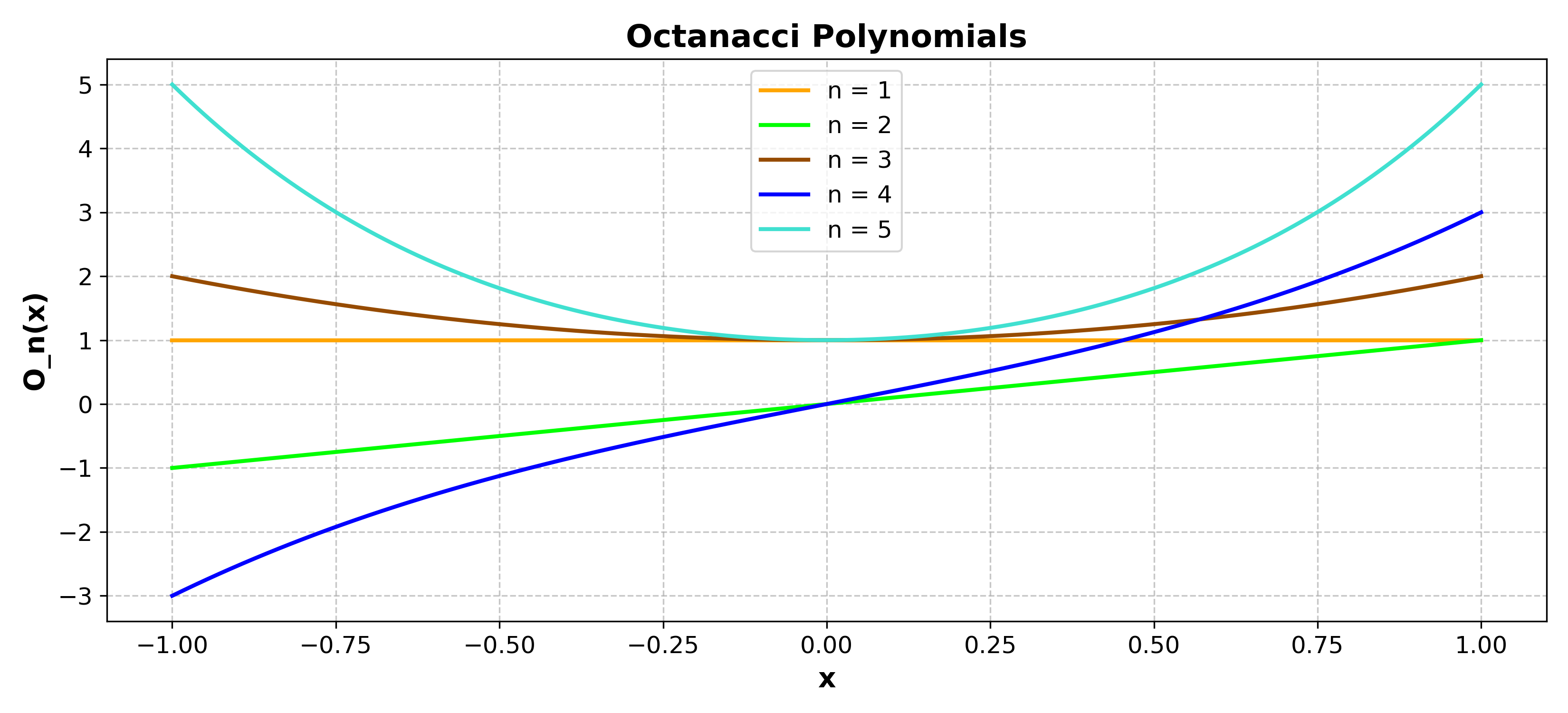}
    \caption{Plot of Octanacci Polynomials for various values of $n$.}
    \label{fig:octanacci_polynomials}
\end{figure}

Figure \ref{fig:octanacci_polynomials} depicts the behavior of Octanacci polynomials for degrees n = 1 to 5. As the degree of the polynomial increases, the oscillations and slopes become more pronounced, particularly in the vicinity of the endpoints of the plotted range. 

\subsubsection{Padé Polynomial}

Padé polynomials, more accurately termed Padé approximants, constitute a family of rational function approximations to a given function. These approximants have notable applications in numerical analysis, particularly in the approximation of special functions and the solution of differential equations\cite{baker1996pade}.
A Padé approximant is the ratio of two polynomials constructed to have the same Taylor series expansion as the function being approximated, up to a given order. For a function $f(x)$, the $[m/n]$ Padé approximant is given by:

\begin{equation}
\label{eq:pade_approximant}
[m/n] = \frac{P_m(x)}{Q_n(x)} = \frac{p_0 + p_1x + p_2x^2 + \cdots + p_mx^m}{1 + q_1x + q_2x^2 + \cdots + q_nx^n}
\end{equation}

where $P_m(x)$ is a polynomial of degree at most $m$, and $Q_n(x)$ is a polynomial of degree at most $n$.

The coefficients of these polynomials are determined by the condition:

\begin{equation}
\label{eq:pade_condition}
f(x) - \frac{P_m(x)}{Q_n(x)} = O(x^{m+n+1})
\end{equation}

This leads to a system of linear equations for the coefficients:

\begin{equation}
\label{eq:pade_system}
\sum_{k=0}^m a_k f_{n+k} = \sum_{k=0}^n b_k f_k
\end{equation}

where $f_k$ are the coefficients of the Taylor series of $f(x)$, and $a_k$, $b_k$ are the coefficients of $P_m(x)$ and $Q_n(x)$ respectively.

\subsubsection{Pentanacci Polynomial}

Pentanacci polynomials represent an intriguing generalization of both the Fibonacci polynomials and the pentanacci numbers \cite{soykan2020generalized}. These polynomials belong to a broader class of generalized Fibonacci polynomials and have significant applications in combinatorics and number theory \cite{kilic2008binet}. The study of Pentanacci polynomials offers valuable insights into higher-order recurrence relations and their polynomial analogues.

The Pentanacci polynomials $P_n(x)$ are defined by the following recurrence relation:

\begin{equation}
\label{eq:pentanacci_recurrence}
P_n(x) = xP_{n-1}(x) + P_{n-2}(x) + P_{n-3}(x) + P_{n-4}(x) + P_{n-5}(x)
\end{equation}

for $n \geq 5$, with initial conditions:
\begin{equation}
\label{eq:pentanacci_initial}
\begin{aligned}
P_0(x) &= 0, \quad P_1(x) = 1, \quad P_2(x) = x, \quad P_3(x) = x^2 + 1, \\
P_4(x) &= x^3 + 2x
\end{aligned}
\end{equation}

\begin{figure}[htbp]
    \centering
    \includegraphics[width=0.8\textwidth]{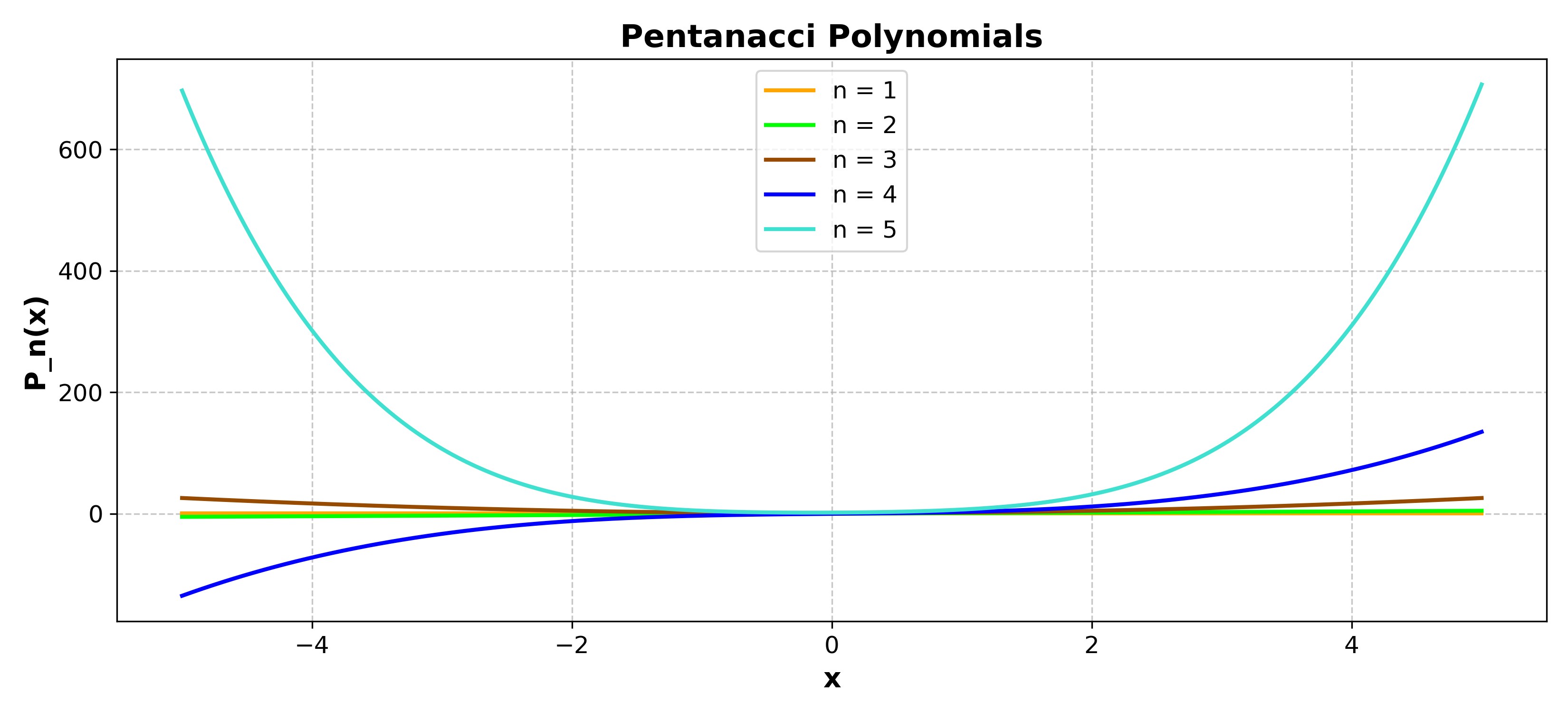}
    \caption{Plot of Pentanacci Polynomials for various values of $n$.}
    \label{fig:pentanacci_polynomials}
\end{figure}

Figure \ref{fig:pentanacci_polynomials} depicts the behavior of Pentanacci polynomials for degrees n = 1 to 5. As the degree of the polynomial increases, the growth rate of the function increases as well, and the oscillations become more pronounced, particularly near the edges of the plotted domain. This highlights the complex characteristics of these distinctive polynomials and how they change with increasing degree.increasing degree.

\subsubsection{Tetranacci Polynomial}
Tetranacci polynomials represent an intriguing generalization of both the Fibonacci polynomials and the tetranacci numbers. These polynomials belong to a broader class of generalized Fibonacci polynomials and have significant applications in combinatorics and number theory \cite{kilic2007tetranacci}. The study of Tetranacci polynomials offers insights into higher-order recurrence relations and their polynomial analogues that are of considerable value.
The Tetranacci polynomials $T_n(x)$ are defined by the following recurrence relation:

\begin{equation}
\label{eq:tetranacci_recurrence}
T_n(x) = xT_{n-1}(x) + T_{n-2}(x) + T_{n-3}(x) + T_{n-4}(x)
\end{equation}

for $n \geq 4$, with initial conditions:
\begin{equation}
\label{eq:tetranacci_initial}
\begin{aligned}
T_0(x) &= 0, \quad T_1(x) = 1, \quad T_2(x) = x, \quad T_3(x) = x^2 + 1
\end{aligned}
\end{equation}

\begin{figure}[htbp]
    \centering
    \includegraphics[width=0.8\textwidth]{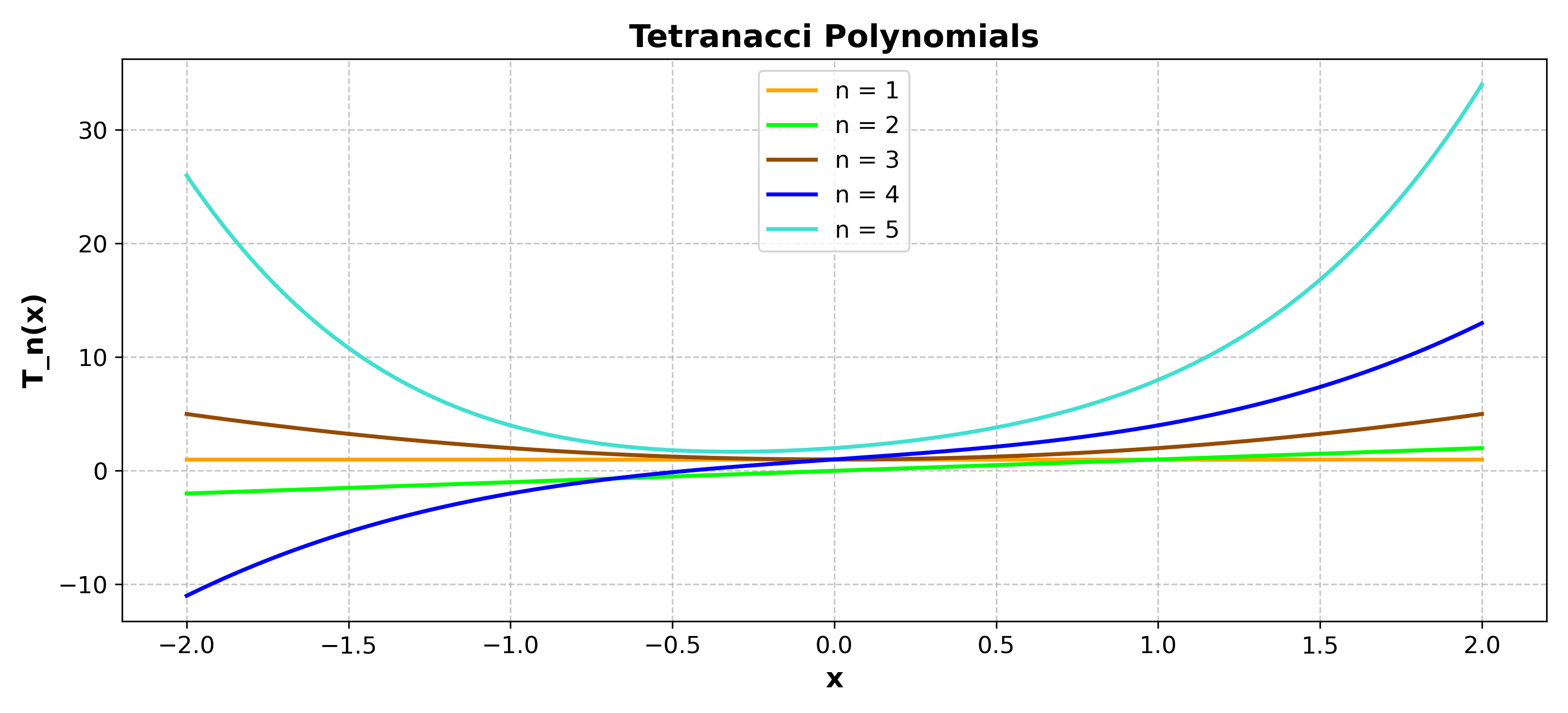}
    \caption{Plot of Tetranacci Polynomials for various values of $n$.}
    \label{fig:tetranacci_polynomials}
\end{figure}

Figure \ref{fig:tetranacci_polynomials} demonstrates the behavior of Tetranacci polynomials for degrees n = 1 to 5. As the degree of the polynomial increases, the growth rate and amplitude of the oscillations also increase, particularly in the vicinity of the plotted range boundaries. This illustrates the complex characteristics of these distinctive polynomials and how their attributes evolve with increasing degree.

\subsubsection{Tribo Polynomial}

Tribo polynomials, also referred to as Tribonacci polynomials, represent an intriguing extension of both the Fibonacci polynomials and the tribonacci numbers \cite{soykan2023generalized}. These polynomials are members of a larger class of generalized Fibonacci polynomials and have notable applications in combinatorics and number theory  \cite{kilic2008combinatorial}. The study of Tribo polynomials provides valuable insights into higher-order recurrence relations and their polynomial analogues.

The Tribo polynomials $T_n(x)$ are defined by the following recurrence relation:

\begin{equation}
\label{eq:tribo_recurrence}
T_n(x) = xT_{n-1}(x) + T_{n-2}(x) + T_{n-3}(x)
\end{equation}

for $n \geq 3$, with initial conditions:
\begin{equation}
\label{eq:tribo_initial}
\begin{aligned}
T_0(x) &= 0, \quad T_1(x) = 1, \quad T_2(x) = x
\end{aligned}
\end{equation}

\begin{figure}[htbp]
\centering
\includegraphics[width=0.8\textwidth]{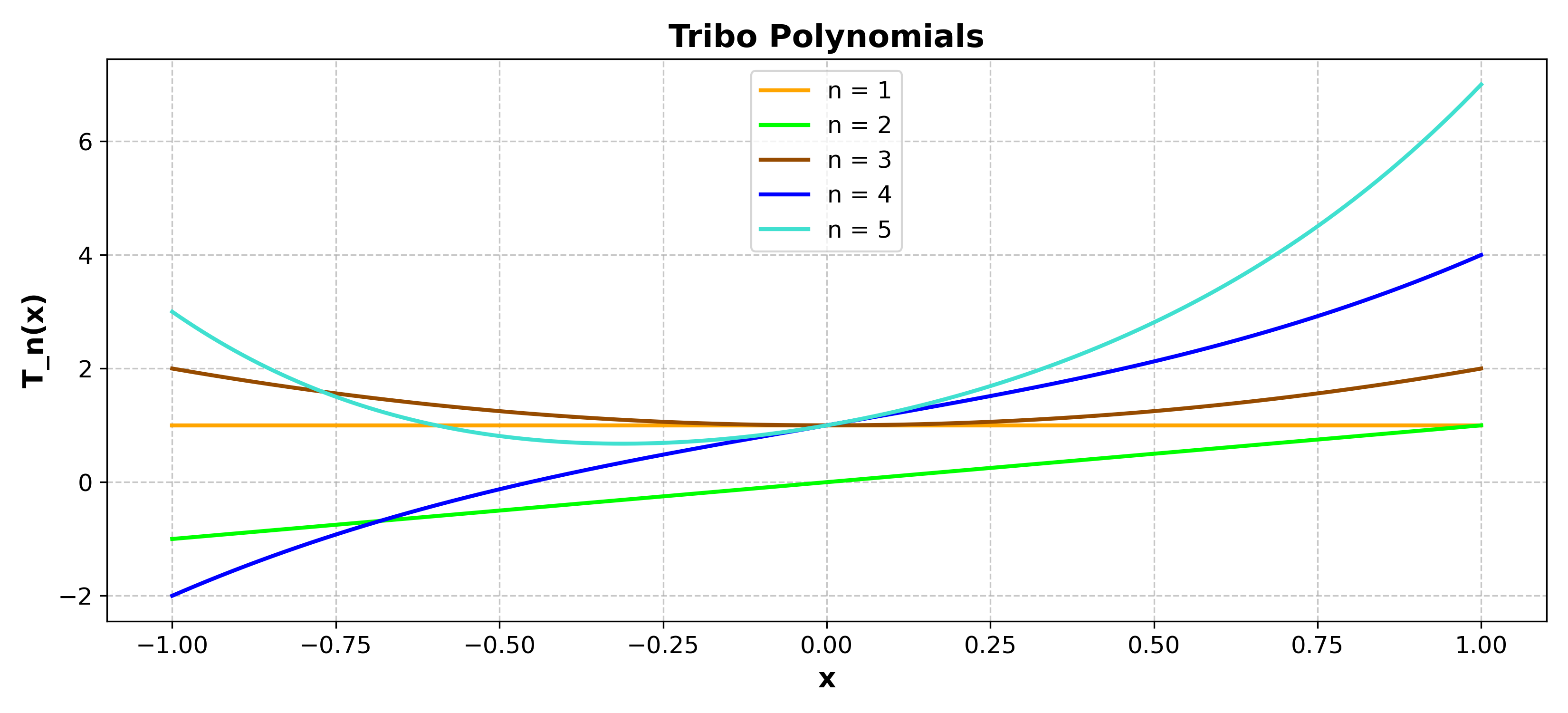}
\caption{Plot of Tribo Polynomials for various values of $n$.}
\label{fig:tribo_polynomials}
\end{figure}

Figure \ref{fig:tribo_polynomials} demonstrates the behavior of tribo polynomials for degrees n = 1 to 5. As the degree of the polynomial increases, the oscillations and slopes of the polynomial exhibit a greater degree of variation, particularly in the vicinity of the endpoints of the plotted interval. This highlights the complex characteristics of these unique polynomials and how they change as the degree increases. 


\subsubsection{VietaPell Polynomial}

Vieta–Pell polynomials represent an intriguing generalization that combines aspects of the Vieta polynomials and the Pell numbers \cite{aziz2021some,witula2006modified}. These polynomials have significant applications in the fields of combinatorics and number theory \cite{shihab2024approximate}. The study of Vieta-Pell polynomials offers insights into the intersection of polynomial sequences and number-theoretic recurrences that are of considerable value.

The VietaPell polynomials $VP_n(x)$ are defined by the following recurrence relation:

\begin{equation}
\label{eq:vietapell_recurrence}
VP_n(x) = 2xVP_{n-1}(x) + VP_{n-2}(x)
\end{equation}

for $n \geq 2$, with initial conditions:
\begin{equation}
\label{eq:vietapell_initial}
\begin{aligned}
VP_0(x) &= 1, \quad VP_1(x) = 2x
\end{aligned}
\end{equation}

\begin{figure}[htbp]
    \centering
    \includegraphics[width=0.8\textwidth]{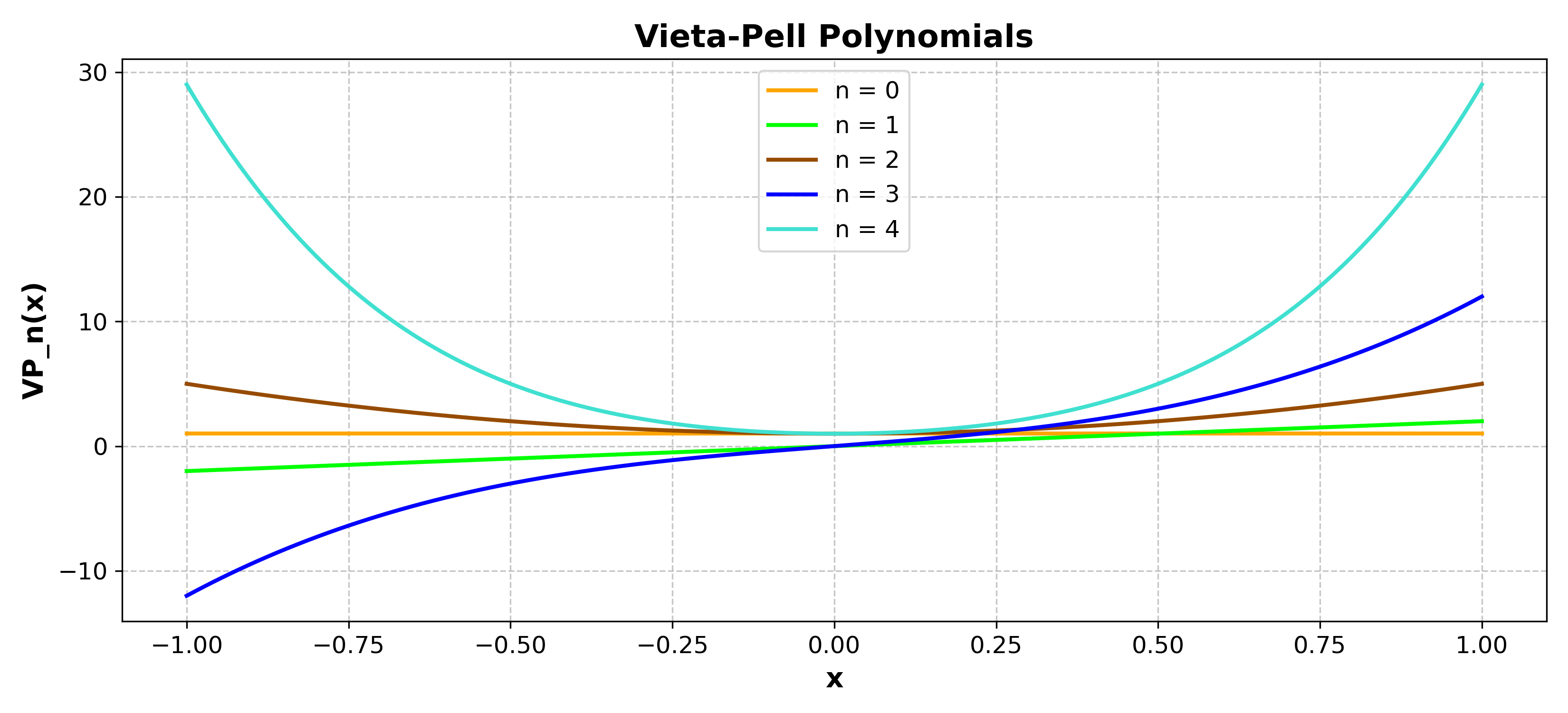}
    \caption{Plot of Vieta–Pell Polynomials for various values of $n$.}
    \label{fig:vietapell_polynomials}
\end{figure}

Figure \ref{fig:vietapell_polynomials} shows the behavior of Vieta–Pell polynomials for degrees n = 0 to 4. As the degree n increases, the polynomials display more pronounced oscillations and faster growth, especially towards the edges of the plotted domain. This highlights the complex nature of these special polynomials.

\subsubsection{Charlier Polynomial}
Charlier polynomials, denoted by $C_n^{(a)}(x)$, constitute a family of discrete orthogonal polynomials that are solutions to the Charlier differential equation\cite{runge1885darstellung, abdul2020computational,zhu2010general}. 
The Charlier polynomials are defined by the generating function:

\begin{equation}
\label{eq:charlier_generating}
\sum_{n=0}^\infty C_n^{(a)}(x) \frac{t^n}{n!} = e^t \left(1 - \frac{t}{a}\right)^x
\end{equation}

where $a > 0$ is a real parameter and $x$ is a non-negative integer.

The explicit formula for Charlier polynomials is given by:

\begin{equation}
\label{eq:charlier_explicit}
C_n^{(a)}(x) = {}_2F_0\left(-n, -x; -; -\frac{1}{a}\right) = \sum_{k=0}^n \binom{n}{k} \frac{(-a)^{n-k}}{k!} x^{\underline{k}}
\end{equation}

where ${}_2F_0$ is a generalized hypergeometric function and $x^{\underline{k}}$ denotes the falling factorial.

\begin{figure}[htbp]
\centering
\includegraphics[width=0.8\textwidth]{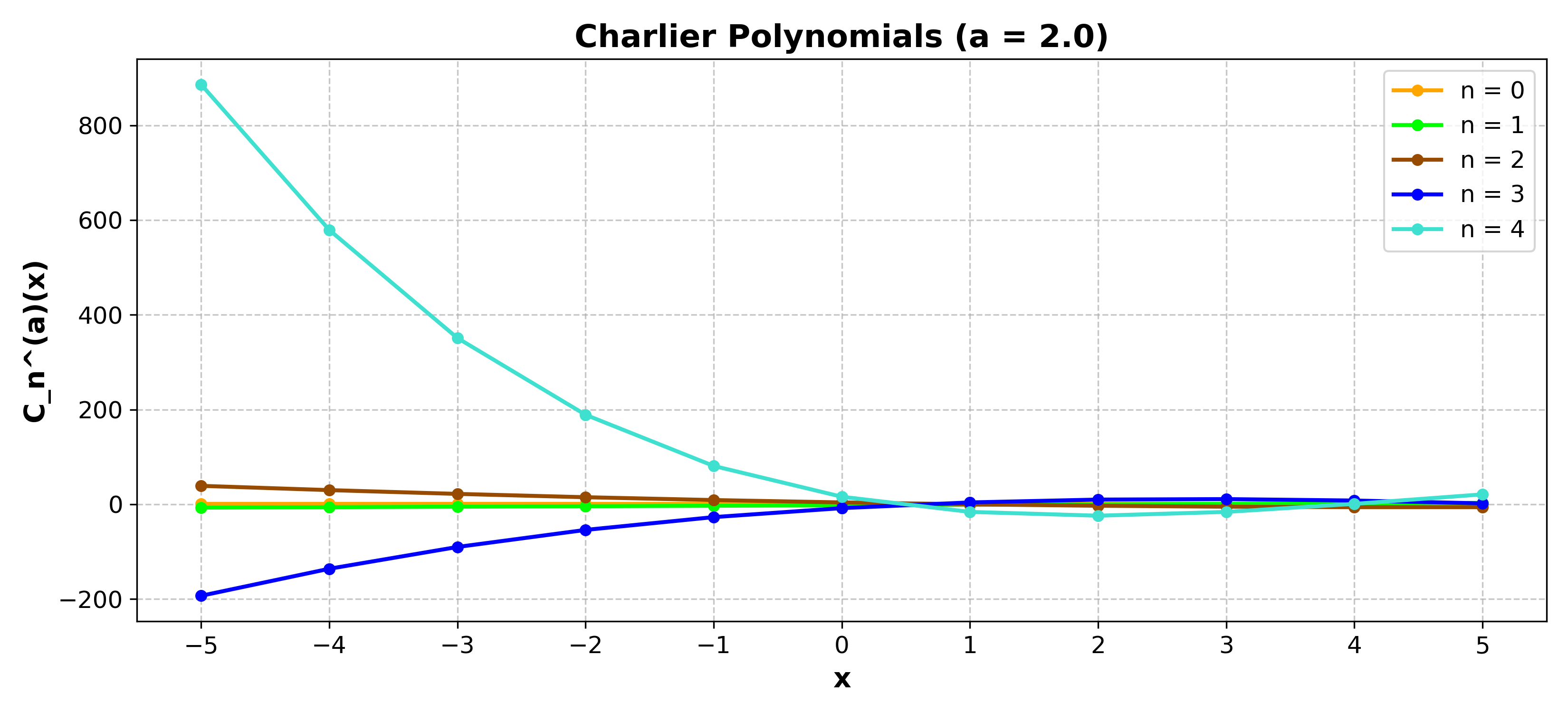}
\caption{Plot of Charlier Polynomials with $a = 2.0$ for various values of $n$.}
\label{fig:charlier_polynomials}
\end{figure}

The Figure \ref{fig:charlier_polynomials} demonstrates the behavior of Charlier polynomials for varying degrees (n = 0 to 4) with parameter a = 2.0. As the degree n increases, the polynomials exhibit heightened oscillations and accelerated growth, particularly at the edges of the plotted range, thereby illustrating the intricate nature of these special functions.

\subsection{Respective Parameter }

Table \ref{tab:polynomial_groups} presents a classification of the 18 polynomials surveyed in the study, along with their respective parameter sets. The polynomials are grouped into several categories based on their mathematical properties and areas of application, such as orthogonal polynomials, hypergeometric polynomials, q-polynomials, Fibonacci-related polynomials, combinatorial polynomials, and number-theoretic polynomials. The parameter column lists the variables or functions that define each polynomial, with some polynomials having no additional parameters beyond their degree.

\begin{table}[h]
\centering
\caption{Classification and parameters of the polynomials surveyed in the study}
\label{tab:polynomial_groups}
\begin{tabular}{|l|l|l|}
\hline
\textbf{Polynomial} & \textbf{Group} & \textbf{Parameters} \\
\hline
Al-Salam-Carlitz & Orthogonal, Hypergeometric, q- & $a$, $q$ \\
\hline
Bannai-Ito & Orthogonal & $\rho_n$, $\tau_n$ \\
\hline
Askey-Wilson & Orthogonal, Hypergeometric, q- & $a$, $b$, $c$, $d$, $q$ \\
\hline
Boas-Buck & Orthogonal & $f(t)$, $g(t)$ \\
\hline
Boubaker & Orthogonal & - \\
\hline
Charlier & Orthogonal & $a$ \\
\hline
Fermat & Number-theoretic & $n$ \\
\hline
Gottlieb & Number-theoretic & - \\
\hline
Heptanacci & Fibonacci-related & - \\
\hline
Hexanacci & Fibonacci-related & - \\
\hline
Meixner-Pollaczek & Orthogonal, Hypergeometric & $\lambda$, $\phi$ \\
\hline
Narayana & Combinatorial & - \\
\hline
Octanacci & Fibonacci-related & - \\
\hline
Padé & Approximation & $m$, $n$, $f_k$ \\
\hline
Pentanacci & Fibonacci-related & - \\
\hline
Tetranacci & Fibonacci-related & - \\
\hline
Tribo & Fibonacci-related & - \\
\hline
Vieta-Pell & Combinatorial, Number-theoretic & - \\
\hline
\end{tabular}
\end{table}

\section{Results}
The table presented in Table~\ref{tab:kan_models} summarizes the performance metrics of various KAN based models, each employing a different type of polynomial as the basis function. These KAN models have been evaluated on the MNIST \cite{mnist}, for handwritten digit classification, and the reported metrics (Overall Accuracy, Kappa, and F1 Score) provide insights into their respective predictive performances. 

By examining the values in the table, the Gottlieb-KAN model exhibits the highest overall accuracy of 0.9759, the highest Kappa of 0.9732, and the highest F1 score of 0.9759, suggesting that it may be the most suitable choice among the listed models for the given task.

\begin{table}[ht]
\centering
\caption{Performance Metrics of KAN Models}
\label{tab:kan_models}
\begin{tabular}{lrrr}
\toprule
Model & Overall Accuracy & Kappa & F1 Score \\
\midrule
Fermat-KAN & 0.9619 & 0.9577 & 0.9619 \\
AlSalam-Carlitz-KAN & 0.9675 & 0.9639 & 0.9675 \\
BannaiIto-KAN & 0.9670 & 0.9633 & 0.9670 \\
Boas-Buck-KAN & 0.9663 & 0.9625 & 0.9663 \\
Boubaker-KAN & 0.9731 & 0.9701 & 0.9731 \\
Charlier-KAN & 0.9726 & 0.9695 & 0.9726 \\
Gottlieb-KAN & \textbf{0.9759} & \textbf{0.9732} & \textbf{0.9759} \\
Heptanacci-KAN & 0.9426 & 0.9362 & 0.9426 \\
Hexanacci-KAN & 0.9641 & 0.9601 & 0.9641 \\
Meixner-Pollacze-KAN & 0.9703 & 0.9670 & 0.9703 \\
Narayana-KAN & 0.9723 & 0.9692 & 0.9723 \\
Octanacci-KAN & 0.9688 & 0.9653 & 0.9688 \\
Pado-KAN & 0.9653 & 0.9614 & 0.9653 \\
Pentanacci-KAN & 0.9542 & 0.9491 & 0.9542 \\
Tetranacci-KAN & 0.9675 & 0.9639 & 0.9675 \\
Tribo-KAN & 0.9569 & 0.9521 & 0.9569 \\
Vieta-Pell-KAN & 0.9749 & 0.9721 & 0.9749 \\
AskeyWilson-KAN & 0.9693 & 0.9659 & 0.9693 \\
\bottomrule
\end{tabular}
\end{table}

An in-depth analysis of the provided data reveals insights into the relationships between model parameters, training time, and overall accuracy. The \textbf{Gottlieb-KAN} model stands out with the highest number of parameters (219,907) and the highest overall accuracy (0.9759), suggesting a potential correlation between model complexity and performance. However, this relationship is not consistent across all models, as some models with similar parameter counts exhibit significant variations in training time and accuracy. The \textbf{Askey-KAN} model, despite having the lowest number of parameters (105,871), exhibits the longest training time (2,418.00 seconds), while the  \textbf{Heptanacci-KAN} model has the shortest training time (729.22 seconds). Factors such as model architecture, optimization techniques, data characteristics, and hardware resources likely contribute to these differences. Furthermore, the overall accuracy values fall within a narrow range (0.9542 to 0.9759), indicating comparable performance across models. While the analysis provides valuable insights, a comprehensive understanding requires further examination of model architectures, training data, optimization strategies, and relevant performance metrics for the specific problem domain. Advanced analytical techniques can quantify the relative importance of various factors and aid in informed model selection and optimization decisions.
\begin{table}[ht]
\centering
\caption{Comparison of the complexity of the model}
\begin{tabular}{lrr}
\toprule
Model & Number of Parameters & Training Time (s) \\
\midrule
AlSalam & 105,862 & 927.30 \\
BannaiIto & 105,865 & 1,261.58 \\
Boas-Buck & 105,880 & 1,220.49 \\
Boubaker & 105,856 & 807.36 \\
Charlier & 105,856 & 853.90 \\
Gottlieb & 219,907 & 923.96 \\
Heptanacci & 105,856 & 729.22 \\
Hexanacci & 105,856 & 760.49 \\
Meixner & 105,862 & 1,095.85 \\
Narayana & 105,856 & 773.70 \\
Octanacci & 105,856 & 770.62 \\
Pado & 105,856 & 764.13 \\
Pentanacci & 105,856 & 762.46 \\
Tetranacci & 105,856 & 767.92 \\
Tribo & 105,856 & 751.97 \\
Vieta-Pell & 105,856 & 792.53 \\
Fermat & 105,856 & 757.21 \\
Askey & 105,871 & 2,418.00 \\
\bottomrule
\end{tabular}
\label{tab:model_details}
\end{table}
\section{Conclusion}
In this study, we conducted a comprehensive survey of 18 polynomials and their potential applications as basis functions in Kolmogorov-Arnold Network (KAN) models. The polynomials were classified into various groups based on their mathematical properties, providing a structured overview of their characteristics and areas of application. The performance of the KAN models incorporating these polynomials was evaluated on the MNIST dataset for handwritten digit classification, with the Gottlieb-KAN model achieving the highest overall accuracy, Kappa, and F1 score.
The analysis of the model complexity, number of parameters, and training time revealed insights into the relationships between these factors and the overall performance. However, the narrow range of accuracy values suggests that further investigation and tuning of these polynomials on more complex datasets are necessary to fully understand their capabilities and potential in KAN models.
This study serves as an initial exploration of the use of various polynomials in KAN models and highlights the need for further research in this area. Future work should focus on applying these polynomials to a wider range of datasets with varying complexity levels and investigating the impact of different model architectures, optimization techniques, and hyperparameter settings on their performance.
Additionally, the development of more advanced analytical techniques to quantify the relative importance of various factors in KAN model performance could aid in informed model selection and optimization decisions. By leveraging the insights gained from this study and conducting further research, we can unlock the full potential of these polynomials in KAN models and advance the field of machine learning.

\bibliographystyle{unsrt}
\bibliography{references}

\end{document}